\documentclass[lettersize,journal]{IEEEtran}
\usepackage{amsmath,amsfonts}
\usepackage{algorithmic}
\usepackage{algorithm}
\usepackage{array}
\usepackage[caption=false,font=normalsize,labelfont=sf,textfont=sf]{subfig}
\usepackage{textcomp}
\usepackage{stfloats}
\usepackage{url}
\newcommand{\eg}[1]{\textit{e.g.,}}
\newcommand{\ie}[1]{\textit{i.e.,}}
\newcommand{\etc}[1]{\textit{etc}}

\usepackage{fontawesome5}

\usepackage[colorlinks]{hyperref}
\usepackage{cite}
\usepackage[utf8]{inputenc} 
\usepackage[T1]{fontenc}    
\usepackage{url}            
\usepackage{booktabs}       
\usepackage{amsfonts}       
\usepackage{nicefrac}       
\usepackage{microtype}      
\usepackage{xcolor}         
\definecolor{ball blue}{rgb}{0.13, 0.67, 0.8}
\definecolor{bleudefrance}{rgb}{0.19, 0.55, 0.91}
\usepackage{bbding}
\usepackage{pifont}
\usepackage{wasysym}
\usepackage{amssymb}

\usepackage{amsmath}
\usepackage{graphicx}
\usepackage{multirow}
\allowdisplaybreaks[4]
\usepackage{amssymb}
\usepackage{array}
\usepackage{enumitem}
\usepackage{color}
\usepackage{colortbl}
\usepackage{bm}
\usepackage{amsmath}
\usepackage{float}
\usepackage{stfloats}
\definecolor{ForestGreen}{rgb}{0.13, 0.55, 0.13}
\newcommand{\stdvu}[1]{\tiny{\color{darkgray}(#1)} {\color{ForestGreen}$\uparrow$}}
\newcommand{\stdvd}[1]{\scriptsize{\color{darkgray}(#1)} {\color{red}$\downarrow$}}

\definecolor{mypink}{rgb}{.99,.91,.95}

\usepackage{wrapfig}

\usepackage{bbding}
\usepackage{pifont}
\usepackage{wasysym}
\usepackage{amssymb}
\usepackage[export]{adjustbox}

\usepackage{placeins}

\definecolor{firebrick}{rgb}{0.7, 0.13, 0.13}
\definecolor{darkpastelgreen}{rgb}{0.01, 0.75, 0.24}
\definecolor{deepskyblue}{rgb}{0.0, 0.75, 1.0}
\definecolor{mypink2}{rgb}{.99,.96,.98}
\definecolor{mypink1}{rgb}{.99,.93,.98}
\definecolor{mypink}{rgb}{.99,.90,.98}
\definecolor{mygray}{rgb}{.95,.95,.95}
\definecolor{lv14}{rgb}{0.5,0.5,0.5}

\definecolor{myblue}{rgb}{255, 255, 255}

\begin{document}

\title{Text to Image for Multi-Label Image Recognition with Joint Prompt-Adapter Learning}

\author{Chun-Mei Feng,~Kai Yu\textsuperscript{\dag},~Xinxing Xu,~Salman Khan,~Rick Siow Mong Goh,\\~Wangmeng Zuo,~\IEEEmembership{Senior~Member,~IEEE},~Yong Liu
        
\thanks{Chun-Mei Feng, Yong Liu, and Rick Siow Mong Goh are with the Institute of High Performance Computing (IHPC), Agency for Science, Technology and Research (A*STAR), Singapore. Email: fengcm.ai@gmail.com}%

\thanks{Kai Yu is with the University of Minnesota, Minneapolis, MN 55455, USA. (email: yu001014@umn.edu).}

\thanks{Xinxing Xu is with the Microsoft Research Asia Singapore.}

\thanks{Salman Khan is with the Mohamed bin Zayed University of Artificial Intelligence (MBZUAI), UAE, and Australian National University, Canberra ACT, Australia.}

\thanks{Wangmeng Zuo is with the Harbin Institute of Technology, Harbin, China. Email: wmzuo@hit.edu.cn.}

\thanks{\textsuperscript{†}Corresponding author.}

}

\markboth{Journal of \LaTeX\ Class Files,~Vol.~14, No.~8, August~2021}%
{Shell \MakeLowercase{\textit{et al.}}: A Sample Article Using IEEEtran.cls for IEEE Journals}

\maketitle

\begin{abstract}
Benefited from image-text contrastive learning, pre-trained vision-language models, \eg, CLIP, allow to direct leverage texts as images (TaI) for parameter-efficient fine-tuning (PEFT). 
While CLIP is capable of making image features to be similar to the corresponding text features, the modality gap remains a nontrivial issue and limits image recognition performance of TaI. 
%
%
Using multi-label image recognition (MLR) as an example, we present a novel method, called T2I-PAL to tackle the modality gap issue when using only text captions for PEFT.  
%
%
The core design of T2I-PAL is to leverage pre-trained text-to-image generation models to generate photo-realistic and diverse images from text captions, thereby reducing the modality gap. 
%
%
%
To further enhance MLR, T2I-PAL incorporates a class-wise heatmap and learnable prototypes. This aggregates local similarities, making the representation of local visual features more robust and informative for multi-label recognition.
For better PEFT, we further combine both prompt tuning and adapter learning to enhance classification performance.
T2I-PAL offers significant advantages: it eliminates the need for fully semantically annotated training images, thereby reducing the manual annotation workload, and it preserves the intrinsic mode of the CLIP model, allowing for seamless integration with any existing CLIP framework.
%
%
Extensive experiments on multiple benchmarks, including MS-COCO, VOC2007, and NUS-WIDE, show that our T2I-PAL can boost recognition performance by $3.47$\% in average above the top-ranked state-of-the-art methods.

\end{abstract}
\begin{IEEEkeywords}
Text to Image, Multi-Label Image Recognition, Prompt Learning, Adapter.
\end{IEEEkeywords}

\section{Introduction}
\IEEEPARstart{I}{n} the recent few years, tremendous progress has been made in large-scale visual-language (VL) pre-trained models~\cite{alayrac2022flamingo}, \eg, CLIP~\cite{radford2021learning}.
Their promising performance has empowered a new learning paradigm for adapting VL pre-trained models to various downstream tasks, \ie, learning adapters or prompts 
in a parameter-efficient manner~\cite{sun2022dualcoop,feng2023diverse,feng2023learning,huang2024learning}. 
In this work, we focus on a specific downstream task, \emph{i.e.}, multi-label image recognition, which requires identifying all semantic labels included in an image~\cite{sun2022dualcoop,chen2019multi,wang2017multi,chen2019learning}.

\begin{figure}[t]
	\begin{center}
		\includegraphics[width=\linewidth]{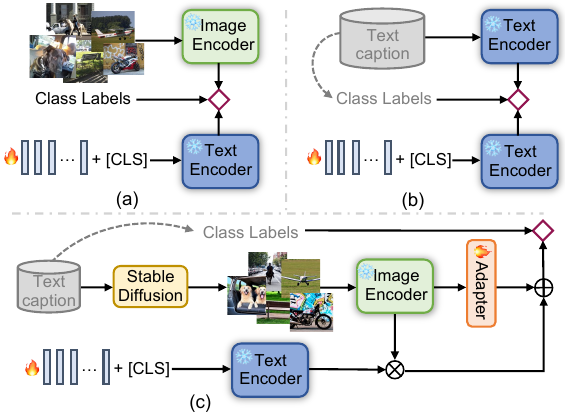}
        \put(-151,148){ \small\rotatebox{270}{$\mathcal{L}_\texttt{Sup}$}}
        \put(-17,148){ \small\rotatebox{270}{$\mathcal{L}_\texttt{Sup}$}}
        \put(-45,84){ \small{$\mathcal{L}_\texttt{Sup}$}}
	\end{center}
	\captionsetup{font=small}
	\caption{\textbf{Comparison} of \textbf{existing prompt tuning} methods and our T2I-PAL for MLR. \textbf{(a)} DualCoOp~\cite{sun2022dualcoop}, which requires a substantial set of annotated images to learn the prompts, thereby being costly in annotation. \textbf{(b)} TaI-DPT~\cite{guo2022texts}, which solely leverages a set of text captions to learn the prompts, but suffers from the modality gap issue. In comparison, we present \textbf{(c)} \textbf{T2I-PAL}, where the pre-trained text-to-image generation model is used to tackle the modality gap, and joint prompt-adapter learning is adopted to improve MLR performance.}
	\label{fig1}
\end{figure}

When adapting VL pre-trained models to MLR, one straightforward method is to annotate full semantic label sets for several images (see Fig.~\ref{fig1} (a)). Nonetheless, exhaustive annotation of MLR gives rise to a much higher cost. 
Fortunately, after large-scale contrastive learning, VL pre-trained models have exhibited promising ability in aligning images with the corresponding text caption. 
Thus, we resort to using text captions as an alternative to images, \ie, TaI-DPT~\cite{guo2022texts}, for learning prompts (see Fig.~\ref{fig1} (b)). 
Contrary to image data, text captions are not only easy to obtain but also explicitly provide the class labels, making them very encouraging for MLR.

%

However, existing VL pre-trained models remain limited in their ability to entirely eliminate the modality gap in the feature space~\cite{gu2022can,nukrai2022text}. 
%
%
Albeit several approaches have been suggested to mitigate this issue~\cite{gu2022can,nukrai2022text}, we present an alternative solution by considering the breakthrough achievements in text-to-image generative models~\cite{nichol2021glide,ramesh2022hierarchical,ruiz2022dreambooth,saharia2022photorealistic}. 
%
In~\cite{gu2022can,nukrai2022text}, noise injection is employed to the textual feature to alleviate the modality gap. 
On the contrary, using text-to-image generation models, one can directly synthesize high-quality and diverse images from text captions. 
Thus, instead of extracting textual features from text captions, extracting image features from synthesized images, is expected to offer a natural solution for MLR.



In this paper, with a set of text captions, we suggest leveraging images synthesized by a text-to-image generation model and joint prompt-adapter, termed T2I-PAL, for MLR, see Fig.~\ref{fig1} (c). T2I-PAL does not require any original training images, nor does it suffer from less performance degradation due to the modality gap caused by using only text captions. To this end, we first crawl captions from public datasets and filter textual descriptions containing one or more target object categories through a noun filter~\cite{guo2022texts}. Then, the text captions containing label information are fed into stable diffusion to obtain synthetic images. T2I-PAL replaces textual features with the image features from synthetic images without modifying the inherent mode of CLIP, thereby circumventing the modality gap of TaI-DPT~\cite{guo2022texts}. To enhance the classification performance, T2I-PAL combines both prompt tuning and adapter learning and simultaneously absorbs the merits of TaI through a shared adapter between text and synthetic images. In particular, T2I-PAL achieves $6.3$\% improvement against TaI-DPT on the {MS-COCO} dataset. To sum up, our contributions are given as follows:

\begin{itemize}[leftmargin=*]
\item
We propose a novel prompt tuning method, termed T2-PAL, which aims to tackle the modality gap issue between text and image when using only text captions for effective MLR.

\item 
T2I-PAL neither requires the original training images with full semantic annotation nor destroys the inherent mode of the CLIP model, allowing it to be embedded into any CLIP model.

\item 
T2I-PAL combines both prompt tuning and adapter learning, and can absorb the benefits of TaI via a shared adapter between text and synthetic images.

\item 
Experimental results show that T2I-PAL significantly outperforms the state-of-the-art MLR methods, and can be combined with existing prompting methods to further improve MLR performance.
\end{itemize}

\section{Related Work}
\vspace{4pt}
\noindent\textbf{Multi-Label Image Recognition.}
MLR aims to train a classifier that can recognize all object categories in an input image~\cite{alfassy2019laso,narayan2021discriminative,simon2022meta}. To establish the correlation between different labels, some works introduce graph neural networks~\cite{chen2019multi,wang2020multi,zxcchen2019learning,zhao2021transformer}, recurrent neural networks~\cite{wang2016cnn,wang2017multi,yazici2020orderless}. 
%
%
Other works adopt object proposals~\cite{wang2016beyond,liu2018multi} or attention mechanisms~\cite{wang2017multi,chen2018recurrent,chen2019learning} as regularization to improve the robustness of the classifier. 
These methods usually require a sufficient number of fully annotated images as training data, and when the data is limited or the labels are limited, the performance of these methods will drop significantly. 
To improve the robustness of MLR, recent works start to focus on the scenarios with few shots or partially labeled data. 
%
%
For example, LaSO~\cite{alfassy2019laso} utilizes the features of paired training images to synthesize samples with multiple labels, allowing the model to promote multi-label few-shot classifier learning in a sample-augmented manner. MS-COCO~\cite{simon2022meta} learns to predict multiple labels in an image by exploring some supporting examples via meta-learning. 
Dat and Sanath \textit{et al.} recognized multiple labels of an image as there is no training signal, \ie, zero-shot, via a multi-attention and a region-based discriminative mechanism~\cite{alfassy2019laso,narayan2021discriminative}. 
For those partial labeled for MLR, where only partial labels are annotated per training image to reduce the annotation cost~\cite{chen2019learning,chen2019multi,durand2019learning,chen2022structured,pu2022semantic}.
%
%
For example, Thibaut \textit{et al.} design the normalization of the partial binary cross-entropy loss to exploit the proportion of known labels per image~\cite{chen2022structured}. 
Tao \textit{et al.} utilize instance- and prototype-level semantic representations to blend category-specific representations in different images to transfer information from known labels to complement unknown labels~\cite{pu2022semantic}.
Inspired by the progress in large VL pre-trained models, DualCoOp leverages the strong alignment of textual and visual features pre-trained by CLIP to learn positive and negative prompts with class names for MLR~\cite{sun2022dualcoop}.
Guo \textit{et al.} alternated the text as images for prompt tuning, revising the default setting of learning prompts for images by visualizing data~\cite{guo2022texts}. Though impressive, the modality gap between the text and image makes it difficult for TaI to adapt to the image domain naturally in the test phase while learning prompts on the text~\cite{guo2022texts}. 
Instead of directly replacing the original image with text, we replace the text captions with synthetic images using the text-to-image generation model. Therefore, we do not need to modify the input schema of the CLIP pre-trained model to make it perform better on downstream tasks.


\vspace{4pt}
\noindent\textbf{Parameter-Efficient Fine-tuning.}
The parameter-efficient fine-tuning (PEFT) mechanism efficiently adapts the pre-trained model to downstream tasks by updating only a small number of model parameters, thereby improving the efficiency of large models such as CLIP and reducing annotation and training costs~\cite{zhu2022prompt,derakhshani2022variational,wang2022learning,mou2023t2i,sung2022vl,zhang2022tip}. 
There are two mainstream methods: adapter-based method, \eg, T2I-Adapter is a lightweight adapter that aligns the internal knowledge in T2I models with external control signals~\cite{mou2023t2i}, VL-Adapter integrates the adapter into the CLIP model and achieves excellent performance in various multi-modal tasks~\cite{sung2022vl}, \ie, image-text and video-text tasks, and Zhang \textit{et al}. proposed Tip-Adapter that does not require training while fine-tuning only the lightweight residual feature adapter, making the performance of few-shot classification be improved~\cite{zhang2022tip}; and prompt-based tuning method that requires a small number of learnable parameters in the input space, such as CoOp~\cite{zhou2022learning} and CoCoOp~\cite{zhou2022conditional} adopting prompt tuning to adapt VL models for downstream image tasks.
%
%
%
%
%
For MLR, DualCoOp~\cite{sun2022dualcoop} and TaI-DPT~\cite{guo2022texts} perform prompt tuning on the VL pre-training model in order to adapt the model to downstream tasks. 
However, in our method, due to the absence of fully annotated image training data, it is difficult to bridge the modality gap between text and images by only adding a small number of new parameters at the input to train multi-label classifiers on synthesized images. 
As a result, we suggest simultaneously introducing prompts and adapters in both input and model, where the adapter shares between the two modalities to explicitly enhance the classification performance on the synthesized image.

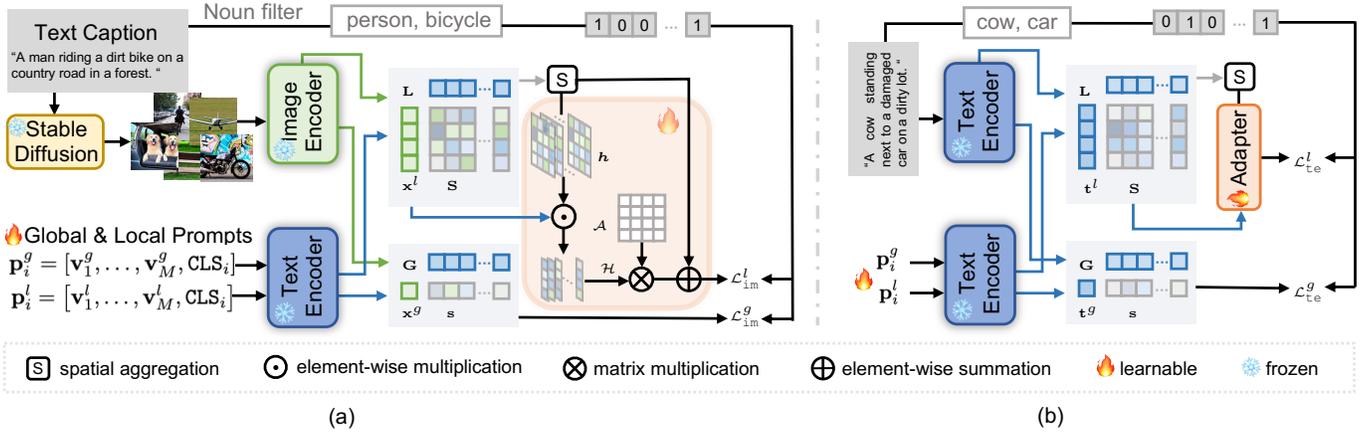
\begin{figure*}[t]
	\begin{center}
		\includegraphics[width=\linewidth]{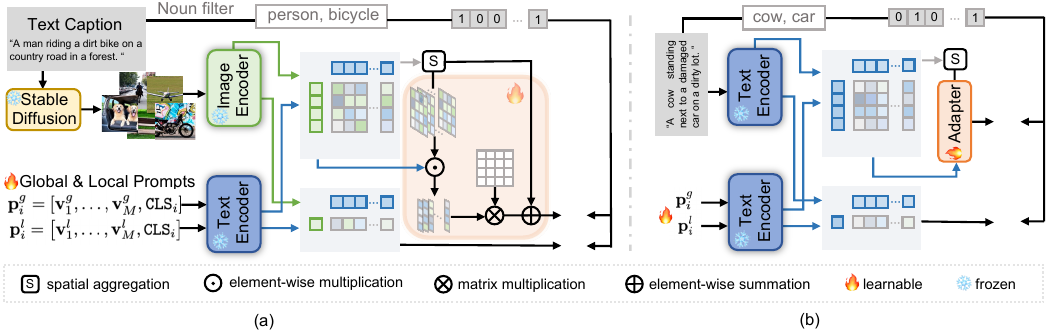}
        \put(-367,128){ \tiny{$\mathbf{L}$}}
        \put(-367,62){ \tiny{$\mathbf{G}$}}
        \put(-111,128){ \tiny{$\mathbf{L}$}}
        \put(-111,62){ \tiny{$\mathbf{G}$}}
        \put(-367,92){ \tiny{$\mathbf{x}^l$}}
        \put(-367,44){ \tiny{$\mathbf{x}^g$}}
        \put(-109,91){ \tiny{$\mathbf{t}^l$}}
        \put(-111,44){ \tiny{$\mathbf{t}^g$}}
        \put(-350,92){ \tiny{$\mathbf{S}$}}
        \put(-350,44){ \tiny{$\mathbf{s}$}}
        \put(-92,91){ \tiny{$\mathbf{S}$}}
        \put(-92,44){ \tiny{$\mathbf{s}$}}
        \put(-293,103){ \tiny{$\boldsymbol{h}_{}$}}
        \put(-292,61){ \tiny{$\mathcal{H}$}}
        \put(-295,77){ \tiny{$\mathcal{A}$}}
        \put(-243,57){ \tiny{$\mathcal{L}_{\texttt{im}}^l$}}
        \put(-243,43){ \tiny{$\mathcal{L}_{\texttt{im}}^g$}}
        \put(-30,101){ \tiny{$\mathcal{L}_{\texttt{te}}^l$}}
        \put(-30,52){ \tiny{$\mathcal{L}_{\texttt{te}}^g$}}
	\end{center}
	\captionsetup{font=small}
	\caption{\textbf{Overall pipeline} of our proposed \textbf{T2I-PAL}. \textbf{(a)} Using pre-trained text-to-image generation models~\cite{rombach2022high} to generate synthesized images from text captions and jointly learning prompt-adapter without modifying the inherent mode of the pertained CLIP. \textbf{(b)} Sharing adapter in the text caption branch to further enhance the classification performance. }
	\label{fig2}
\end{figure*}

\vspace{4pt}
\noindent\textbf{Synthetic Data for Image Recognition.}
The synthesized image has shown excellent performance in an increasing number of tasks due to its high flexibility~\cite{choi2020stargan,rombach2022high,nichol2021glide,sinha2021d2c,ho2022imagen}. Using early image generation methods, \eg, VAEs~\cite{kingma2013auto} and GANs~\cite{goodfellow2020generative}, initial attempts have been made in some visual tasks. In recent years, diffusion models have gradually become promising and powerful generative models that perform well in many applications, \eg, high-resolution image synthesis~\cite{rombach2022high,liu2022compositional}, text-to-image generation~\cite{nichol2021glide,ramesh2022hierarchical}, few-shot conditional image generation~\cite{sinha2021d2c}, as well as point cloud generation~\cite{ho2022imagen}. Several diffusion-based text-to-image models, including Stable Diffusion~\cite{rombach2022high}, DALL-E2~\cite{ramesh2022hierarchical}, Imagen~\cite{saharia2022photorealistic}, and GLIDE~\cite{nichol2021glide}, have been developed, providing an unprecedented synthesis quality and promoting the development of the AI-for-Art community. 
In particular, text-to-image generation can be seen as a conditional image generation task, which only requires inputting some natural language descriptions of what we want to express in the image and outputting it visually. This motivates us to leverage images synthesized by text-to-image generation model for training MLR using only text captions. 
%

\section{Method}\label{method}
\label{headings}
\vspace{4pt}
In order to train a multi-label classifier by exploiting the large-scale pre-training models (CLIP~\cite{radford2021learning}) in the absence of training images, we advocate leveraging synthesized images from text to image (T2I).
With synthesized images, we do not need to modify the input modality of CLIP and thus can inherit the merits of large-scale visual language pre-training. Contrarily, treating text as an image will reduce the performance of the CLIP model in downstream tasks due to the modality gap between text and image.
We note that the discussion on the modality gap between original and synthetic images is provided in Section~\ref{ab3}, where the results show that there is no significant difference between the two types of data, which motivates us to use synthetic images in our work.
As shown in Fig.~\ref{fig2} (a), T2I-PAL first constructs text captions from public image caption datasets~\cite{lin2014microsoft,krasin2017openimages} as input to T2I for image synthesis~\cite{guo2022texts}. After that, we feed these synthesized images and the constructed text description into the image and text encoders, respectively, and freeze them. During training, we adopt three encoders from the pre-trained CLIP, \ie, one image encoder to encode the synthesized image, two text encoders for the prompts and text captions. Additionally, to enhance the classification performance, T2I-PAL combines both prompt tuning and adapter learning, and can absorb the benefits of TaI via a shared adapter between text and synthetic images.
%
%
During testing, the learned prompts are encoded by a text encoder to output the class embeddings, see Fig.~\ref{fig2} (b). The other text and visual encoders are replaced by a visual encoder and learned adapter, which receives test images as input and extracts the features of each test image. These features are combined with global and local prompts to generate class embeddings through cosine similarity to give the final classification result.



\subsection{Preparation of Captions to Synthesize Images.}
Following~\cite{guo2022texts}, we extract captions from public image captioning datasets (\eg, MS-COCO~\cite{lin2014microsoft}) and localized narratives from object detection datasets (\eg, OpenImages~\cite{krasin2017openimages}) for generating the synthetic image. 
Among them, the labels used for training are also extracted from the caption, enabling no information from pictures to be disclosed during training~\cite{guo2022texts}. 
Concretely, given a multi-label dataset $\mathcal{X}$ with a set of object categories $\mathcal{C}=\left\{\mathrm{c}_1, \mathrm{~c}_2, \mathrm{~c}_3, \ldots, \mathrm{c}_C\right\}$, where $C$ is the number of categories, we first map nouns with similar meanings to their corresponding category labels through the noun filter, and then search for sentences containing at least one class name $\mathrm{c}_i$ in $\mathcal{C}$, otherwise remove the irrelevant sentences directly. 
With the constructed text description, we feed it into a pre-trained T2I generation model, Stable Diffusion~\cite{rombach2022high}, to synthesize its corresponding visual data $\mathcal{X}^{'}$, refer to the Fig.~\ref{fig2} (a) for details. 
In this way, we can use the paired labels and synthetic data to finetune the pre-trained model, \ie, CLIP~\cite{radford2021learning}, in a parameter-efficient manner.

\subsection{T2I for Joint Prompt-Adapter Learning.}\label{adapter}
Following~\cite{guo2022texts}, to identify all labeled objects in the MLR task, global and local prompts are added to the class tokens to achieve discrimination by the coarsest-grained and fine-grained features, respectively. Formally, we have
\begin{equation}
\begin{aligned}
\mathbf{p}_i^g =\left[\mathbf{v}_1^{g}, \mathbf{v}_2^{g}, \mathbf{v}_3^{g}, \ldots, \mathbf{v}_M^{g}, \texttt{CLS}_{i}\right], \\
\mathbf{p}_i^l =\left[\mathbf{v}_1^{l}, \mathbf{v}_2^{l}, \mathbf{v}_3^{l}, \ldots, \mathbf{v}_M^{l}, \texttt{CLS}_{i}\right],
\end{aligned}
\end{equation}
where $\mathbf{p}^g$ and $\mathbf{p}^l$ are the global and local prompts consist with the learnable embedding 
$\mathbf{v}_j^g$, $\mathbf{v}_j^{l}, j \in\{1, \ldots, M\}$ and class tokens \texttt{CLS}$_i$ with $i$-th class in $C$ categories. 
%
$M$ denotes the number of learnable embeddings in both the global and local prompts, defined by specifying the number of learnable vectors $\mathbf{v}_j^g$, $\mathbf{v}_j^{l}$ they contain.
As shown in Fig.~\ref{fig2}, we then use the pre-trained CLIP text encoder $\operatorname{E}_\texttt{{te}}$ to generate the global $\mathbf{G}=\left\{\mathbf{G}_i\right\}_{i=1}^C$ and local $\mathbf{L}=\left\{\mathbf{L}_i\right\}_{i=1}^C$ class embeddings. Accordingly, we extract global $\mathbf{x}^g \in \mathbb{R}^{N_\texttt{im} \times D}$ and local $\mathbf{x}^l\in \mathbb{R}^{N_\texttt{te} \times D}$ visual features from the pre-trained CLIP visual encoder $\operatorname{E}_\texttt{{im}}$ with a visual input of synthetic image $\mathbf{x}$, where the $\mathbf{x}^l$ is the feature map before the attention pooling layer of CLIP. $N_\texttt{im}$ and $N_\texttt{te}$ are the image size and length of text tokens, respectively. Analogously, we copy a text encoder $\operatorname{E}_\texttt{{te}}$ of CLIP with a textual input of a piece of training text description to generate the global $\mathbf{t}^g\in \mathbb{R}^{D}$ and local $\mathbf{t}^l\in \mathbb{R}^{D}$ text features.

Given this, the global and local similarities of both visual and text features can be obtained by
\begin{equation}
\mathbf{s}_i = \left\langle \boldsymbol{f}^g, \mathbf{G}_i\right\rangle, \ \mathbf{S}_{ij} = \left\langle \boldsymbol{f}^l_j, \mathbf{L}_i\right\rangle,
\label{eq2}
\end{equation}
where $\boldsymbol{f}^g$ and $\boldsymbol{f}^l$ indicate the global (\ie, $\mathbf{x}^g$ or $\mathbf{t}^g$) and local (\ie, $\mathbf{x}^l$ or $\mathbf{t}^l$) features from the pre-trained CLIP encoder.
We then aggregate the local similarities in a spatially weighted manner
\begin{equation}
\mathbf{s}_i^{\prime}=\sum_{j=1}^{N_{*}} \frac{\exp \left(\mathbf{S}_{i j} / \tau\right)}{\sum_{j=1}^{N_{*}} \exp \left(\mathbf{S}_{i j} / \tau\right)} \cdot \mathbf{S}_{i j},
\label{eq3}
\end{equation}
where ${N_{*}}$ is the size of the image or length of text tokens, $\tau$ refers to the ability to focus on a specific location.

For enhancing classification performance, we further combine both prompt tuning and adapter learning. In particular,
%
%
Zhang \textit{et al}. proposed the Tip-Adapter~\cite{zhang2022tip} by constructing query-key cache model from few-shot supervisions that can provide better vision-language modeling. With this perspective, we establish an adapter with query-key pairs, \ie, local features treated as query, an initialized matrix with a size of $D\times C$ treated as key, that can be shared with both the visual and text features.

As shown in Fig.~\ref{fig2}, the local text ($\mathbf{t}^l$) and visual features ($\mathbf{x}^l$), extracted by $\operatorname{E}_\texttt{{te}}$ and $\operatorname{E}_\texttt{{im}}$, serve as queries for retrieval from the learnable matrix $\mathcal{A}\in \mathbb{R}^{ C \times D}$. Note that $\mathcal{A}$ is shared between the text and image branches.
To enhance the network's ability to recognize multiple classes, we introduce a method that leverages a class-wise heat map to enrich the representation of local visual features $\boldsymbol{f}_j^l$ across different classes. 
Given the local class embedding $\mathbf{L}_i$, we calculate the local similarity $\mathbf{S}_{i j}$ as their inner product. Then the class-wise heat map can be derived as
\begin{equation}
\boldsymbol{h}_{ij}=\frac{\exp \left(\mathbf{S}_{i j} / \tau\right)}{\sum_{j=1}^{N_{*}} \exp \left(\mathbf{S}_{i j} / \tau\right)}.
\end{equation}
Class-wise heat map contributes to enhancing classification performance from two aspects:
(1) Local similarities are fragile and noisy. Class-wise heat map can be used to aggregate the local similarities to obtain a robust class-wise similarity $\mathbf{s}_i^{\prime}=\sum_{j=1}^{N *} \boldsymbol{h}_{i j} \cdot \mathbf{S}_{i j}$ (see Eq.~\ref{eq3}).
(2) Moreover, class-wise heat map can be used to obtain the class-wise attended feature 
$\mathcal{H}_i=\sum_{j=1}^{N_*} \boldsymbol{h}_{i j} \cdot \boldsymbol{f}_j^l$. Let $\mathcal{H}=\left[\mathcal{H}_1, \ldots, \mathcal{H}_i, \ldots, \mathcal{H}_C\right]$. 
We further introduce a learnable prototype matrix $\mathcal{A}=\left[\mathcal{A}_1, \ldots, \mathcal{A}_i, \ldots, \mathcal{A}_C\right]$, where $\mathcal{A}_i$ is the learnable prototype of class $i$.
Consequently, we use
\begin{equation}
\boldsymbol{q}=\operatorname{diag} \left( \exp \left(-\beta\left(1-\mathcal{H} \mathcal{A}^T \right)\right) \right),
\label{eq6}
\end{equation}
to denote the affinity to the $i$-th class prototype, where $\beta$ refers the modulating hyperparameter~\cite{zhang2022tip}. 
As noted, $\boldsymbol{f}^l$ can be either local CLIP text feature $\boldsymbol{t}^l$ or image feature $\boldsymbol{x}^l$. 
For $\boldsymbol{t}^l$ either or $\boldsymbol{x}^l$, we can use Eq.~\ref{eq6} to compute the class-wise affinity between $\mathcal{H}_i$ and $\mathcal{A}_i$. 
In terms of shared $\mathcal{A}$, we mean that the same $\mathcal{A}$ is adopted for both $\boldsymbol{t}^l$ and $\boldsymbol{x}^l$.
In this way, both $\boldsymbol{t}^l$ and $\boldsymbol{x}^l$ contribute to better training of the learnable prototype matrix $\mathcal{A}$, thereby enhancing the class representation ability.
Taking both $\mathbf{s}_i^{\prime}$ and $\boldsymbol{q}_i$ into account, we define local predicted logits of the joint prompt-adapter as
\begin{equation}
\begin{aligned} \widetilde{\mathbf{s}}_i^{\prime} & =\alpha \boldsymbol{q}_i + \mathbf{s}_i^{\prime},
\end{aligned}
\label{eq7}
\end{equation}
where $\alpha$ is the residual ratio of the features of the CLIP’s text or visual encoder. 
To sum up, class-wise heat map contributes to both local prompt (\ie, $\mathbf{s}_i^{\prime}$) and adapter (\ie, $\boldsymbol{q}_i$).
That is, T2I-PAL receives not only the prior knowledge of the pre-trained CLIP's visual encoder but also the new knowledge that the adapter collects from the text. Accordingly, the smaller the value of $\alpha$, the more prior knowledge needs to be acquired from the pre-trained CLIP's visual encoder, while the larger the value of $\alpha$, the more knowledge needs to be learned from the adapter. More importantly, as the learnable matrix $\mathcal{A}$ is shared with the synthetic image and text caption branches, it encourages the synthetic image to absorb the merits of the text caption, thereby enhancing the class representation ability of the classifier.

\subsection{Learning Objective.} 
The overall loss of our proposed T2I-PAL can be expressed as $\mathcal{L} = \gamma \mathcal{L}_{\texttt{im}} + (1-\gamma)\mathcal{L}_{\texttt{te}} $, where $\gamma$ is the trade-off weighted the two terms. Concretely, both the $\mathcal{L}_{\texttt{im}}$ and $\mathcal{L}_{\texttt{te}}$ contain two terms, \ie, the global and local similarities in Eq.~\ref{eq2}. Following~\cite{guo2022texts}, ranking loss~\cite{gong2013deep} is adopted to measure the discrepancy between the classification score learned from a synthetic image and text caption with the ground-truth labels. The details can be formulated as
\begin{equation}
\begin{aligned}
& \mathcal{L}_{\texttt{*}}^g=\sum\nolimits_{i \in\left\{c^{+}\right\}} \sum\nolimits_{j \in\left\{c^{-}\right\}} \max \left(0, \eta-\left\|\mathbf{s}_i-\mathbf{s}_j\right\|_2\right), \\
& \mathcal{L}_{\texttt{*}}^l=\sum\nolimits_{i \in\left\{c^{+}\right\}} \sum\nolimits_{j \in\left\{c^{-}\right\}} \max \left(0, \eta-\left\|\widetilde{\mathbf{s}}_i^{\prime}-\widetilde{\mathbf{s}}_j^{\prime}\right\|_2\right),
\end{aligned}
\end{equation}
where $\eta$ refers the margin value that determines the minimum amount by which the similarity score between positive classes should be greater than that of negative classes~\cite{guo2022texts}.

\section{Experiments}

\subsection{Experimental Setup}
\vspace{4pt}
\noindent\textbf{Implementation Details.} We implemented our method with Pytorch on one NVIDIA Tesla A$100$ GPU with $40$GB of memory. The visual and text encoders are initialized from the CLIP pre-trained model with ResNet-$101$ and Transformer, respectively.
The impact of different visual encoders, \ie, ResNet-$50$ and ResNet-$101$, on model performance is also investigated in Section~\ref{ab3}.
During training, the pre-trained encoder and decoder are frozen while only the prompts and adapters are optimized by the SGD~\cite{kingma2014adam,loshchilov2017decoupled} with an epoch of $40$ for all datasets. Concretely, the class-specific prompting is initialized with a Gaussian noise sampled from $\mathcal{N}(0,0.02)$, where the length of both the global and local prompts is with a size of $16$. We initialized the adapter with a size of $512$ $\times$ $C$, where $C$ is the number of categories. The batch sizes are set to $64$ and learning rates are initialized at $1$e-$4$ for all three datasets. The hyperparameters of $\gamma$, $\alpha$, $\beta$, $\eta$, and $\tau_s$ are empirically set to $0.2$, $1$, $3.5$, $1$, and $0.02$, respectively.

\begin{table}[t]
\renewcommand{\arraystretch}{1.3}
	\caption{\small \textbf{Comparison} with state-of-the-arts under \texttt{zero-shot} setting, where ${\color{ForestGreen}\uparrow}$ indicates \textbf{improvements} compared with the \textbf{top-1} ranked baseline method, \ie, TaI-DPT.}
	\label{tab1}
        \setlength{\tabcolsep}{2.6pt}
	\fontsize{9}{9}\selectfont
	\centering
	\begin{tabular}{ l  cc cc cc}
\toprule
Method
&\multicolumn{2}{c}{MS-COCO}
&\multicolumn{2}{c}{{VOC 2007}} 
&\multicolumn{2}{c}{NUS-WIDE}\\ 

\cmidrule(r){1-1} \cmidrule(r){2-3}  \cmidrule(r){4-5}  \cmidrule(r){6-7}  

{ZSCLIP~\cite{radford2021learning}~${\color{gray}_{\text{[ICML}21]}}$}
&\multicolumn{2}{|c}{$47.3$} 
&\multicolumn{2}{c}{$76.2$}
&\multicolumn{2}{c}{$36.4$}  \\

{TaI-DPT~\cite{guo2022texts}~${\color{gray}_{\text{[CVPR}23]}}$}
&\multicolumn{2}{|c}{$65.1$} 
&\multicolumn{2}{c}{$88.3$}
&\multicolumn{2}{c}{$46.5$}  \\ 

{PVP~\cite{wu2024tai}~${\color{gray}_{\text{[arXiv}24]}}$}
&\multicolumn{2}{|c}{$70.8$} 
&\multicolumn{2}{c}{$90.0$}
&\multicolumn{2}{c}{$46.0$}
\\ 

{Dong \textit{et al}.~\cite{kang2024class}~${\color{gray}_{\text{[OpenReview}24]}}$}
&\multicolumn{2}{|c}{$70.0$}
&\multicolumn{2}{c}{$89.2$}
&\multicolumn{2}{c}{$46.6$}
\\ 

\cmidrule(r){1-1} \cmidrule(r){2-7}  

\texttt{T2I-PAL+Llama}
&\multicolumn{2}{|c}{$71.2$}
&\multicolumn{2}{c}{$91.4$}
&\multicolumn{2}{c}{$47.3$}
\\ 

{\cellcolor{myblue}$\textbf{{\texttt{T2I-PAL(Ours)}}}$} 
&\multicolumn{2}{|c}{{\cellcolor{myblue}\textbf{71.4}\stdvu{$\underline{{0.6}}$}}} 
&\multicolumn{2}{c}{{\cellcolor{myblue}\textbf{91.5}\stdvu{$\underline{{1.5}}$}}}
&\multicolumn{2}{c}{{\cellcolor{myblue}\textbf{47.4}\stdvu{$\underline{{0.8}}$}}}\\
\bottomrule
\end{tabular}
\end{table}

\FloatBarrier
\begin{table*}[h]
\renewcommand{\arraystretch}{1.3}
	\caption{\small \textbf{Comparison} with state-of-the-arts under \texttt{partial-label} setting, where +TaI-DPT~\cite{guo2022texts} and +T2I-PAL indicate integration with the MLR method, DualCoOp~\cite{sun2022dualcoop}, respectively.} 
	\label{tab2}
        \setlength{\tabcolsep}{9pt}
	\fontsize{9}{9}\selectfont
	\centering
	\begin{tabular}{l l  cc cc cc cc cc cc cc cc cc cc}
\toprule
&Method
&\multicolumn{2}{c}{\textbf{$10$\%}}
&\multicolumn{2}{c}{\textbf{$20$\%}}
&\multicolumn{2}{c}{\textbf{$30$\%}}
&\multicolumn{2}{c}{\textbf{$40$\%}}
&\multicolumn{2}{c}{\textbf{$50$\%}}
&\multicolumn{2}{c}{\textbf{$60$\%}}
&\multicolumn{2}{c}{\textbf{$70$\%}}
&\multicolumn{2}{c}{\textbf{$80$\%}}
&\multicolumn{2}{c}{\textbf{$90$\%}}
&\multicolumn{2}{c}{\textbf{Avg.}}
\\ 

\cmidrule(r){2-2}  
\cmidrule(r){3-20} 
\cmidrule(r){21-22} 

\multirow{7}{*}{\rotatebox{90}{\textbf{MS-COCO}}}

&\multicolumn{1}{|l}{SSGRL~\cite{chen2019learning}~${\color{gray}_{\text{[ICCV}19]}}$}
&\multicolumn{2}{|c}{$62.5$} 
&\multicolumn{2}{c}{$70.5$}
&\multicolumn{2}{c}{$73.2$}  
&\multicolumn{2}{c}{$74.5$}
&\multicolumn{2}{c}{$76.3$}  
&\multicolumn{2}{c}{$76.5$}
&\multicolumn{2}{c}{$77.1$}  
&\multicolumn{2}{c}{$77.9$}
&\multicolumn{2}{c}{$78.4$}  
&\multicolumn{2}{|c}{$74.1$}  
\\

&\multicolumn{1}{|l}{GCN-ML~\cite{chen2019multi}~${\color{gray}_{\text{[CVPR}19]}}$}
&\multicolumn{2}{|c}{$63.8$} 
&\multicolumn{2}{c}{$70.9$}
&\multicolumn{2}{c}{$72.8$}  
&\multicolumn{2}{c}{$74.0$}
&\multicolumn{2}{c}{$76.7$}  
&\multicolumn{2}{c}{$77.1$}
&\multicolumn{2}{c}{$77.3$}  
&\multicolumn{2}{c}{$78.3$}
&\multicolumn{2}{c}{$78.6$}  
&\multicolumn{2}{|c}{$74.4$}  
\\

&\multicolumn{1}{|l}{Par.BCE~\cite{durand2019learning}~${\color{gray}_{\text{[CVPR}19]}}$}
&\multicolumn{2}{|c}{$61.6$} 
&\multicolumn{2}{c}{$70.5$}
&\multicolumn{2}{c}{$74.1$}  
&\multicolumn{2}{c}{$76.3$}
&\multicolumn{2}{c}{$77.2$}  
&\multicolumn{2}{c}{$77.7$}
&\multicolumn{2}{c}{$78.2$}  
&\multicolumn{2}{c}{$78.4$}
&\multicolumn{2}{c}{$78.5$}  
&\multicolumn{2}{|c}{$74.7$}  
\\

&\multicolumn{1}{|l}{SARB~\cite{pu2022semantic}~${\color{gray}_{\text{[AAAI}22]}}$}
&\multicolumn{2}{|c}{$71.2$} 
&\multicolumn{2}{c}{$75.0$}
&\multicolumn{2}{c}{$77.1$}  
&\multicolumn{2}{c}{$78.3$}
&\multicolumn{2}{c}{$78.9$}  
&\multicolumn{2}{c}{$79.6$}
&\multicolumn{2}{c}{$79.8$}  
&\multicolumn{2}{c}{$80.5$}
&\multicolumn{2}{c}{$80.5$}  
&\multicolumn{2}{|c}{$77.9$}  
\\

&\multicolumn{1}{|l}{DualCoOp~\cite{sun2022dualcoop}~${\color{gray}_{\text{[NeurIPS}22]}}$}
&\multicolumn{2}{|c}{$78.7$} 
&\multicolumn{2}{c}{$80.9$}
&\multicolumn{2}{c}{$81.7$}  
&\multicolumn{2}{c}{$82.0$}
&\multicolumn{2}{c}{$82.5$}  
&\multicolumn{2}{c}{$82.7$}
&\multicolumn{2}{c}{$82.8$}  
&\multicolumn{2}{c}{$83.0$}
&\multicolumn{2}{c}{$83.1$}  
&\multicolumn{2}{|c}{$81.9$}  
\\

&\multicolumn{1}{|l}{+TaI-DPT~\cite{guo2022texts}~${\color{gray}_{\text{[CVPR}23]}}$}
&\multicolumn{2}{|c}{$81.5$} 
&\multicolumn{2}{c}{$82.6$}
&\multicolumn{2}{c}{$83.3$}  
&\multicolumn{2}{c}{$83.7$}
&\multicolumn{2}{c}{$83.9$}  
&\multicolumn{2}{c}{$84.0$}
&\multicolumn{2}{c}{$84.2$}  
&\multicolumn{2}{c}{$84.4$}
&\multicolumn{2}{c}{$84.5$}
&\multicolumn{2}{|c}{$83.6$}  \\ 

&\multicolumn{1}{|l}{Positive CoOp}~\cite{rawlekar2024rethinking}~${\color{gray}_{\text{[arXiv}24]}}$
&\multicolumn{2}{|c}{$79.8$}
&\multicolumn{2}{c}{$82.1$}
&\multicolumn{2}{c}{$83.0$}  
&\multicolumn{2}{c}{$83.5$}
&\multicolumn{2}{c}{$83.7$}  
&\multicolumn{2}{c}{$83.9$}
&\multicolumn{2}{c}{$84.0$} 
&\multicolumn{2}{c}{$84.2$}
&\multicolumn{2}{c}{$84.4$}
&\multicolumn{2}{|c}{$83.2$}  \\

&\multicolumn{1}{|l}{PVP}~\cite{wu2024tai}~${\color{gray}_{\text{[arXiv}24]}}$
&\multicolumn{2}{|c}{$81.8$} 
&\multicolumn{2}{c}{$82.8$}
&\multicolumn{2}{c}{$83.3$}  
&\multicolumn{2}{c}{$83.6$}
&\multicolumn{2}{c}{$83.9$} 
&\multicolumn{2}{c}{$84.1$}
&\multicolumn{2}{c}{$84.3$} 
&\multicolumn{2}{c}{$84.6$}
&\multicolumn{2}{c}{$84.8$}
&\multicolumn{2}{|c}{$83.7$} \\ 

&\multicolumn{1}{|l}{Dong \textit{et al}.}~\cite{kang2024class}~${\color{gray}_{\text{[OpenReview}24]}}$
&\multicolumn{2}{|c}{$81.5$}
&\multicolumn{2}{c}{$82.8$}
&\multicolumn{2}{c}{$83.3$} 
&\multicolumn{2}{c}{$83.5$}
&\multicolumn{2}{c}{$84.0$}  
&\multicolumn{2}{c}{$84.2$}
&\multicolumn{2}{c}{$84.4$} 
&\multicolumn{2}{c}{$84.5$}
&\multicolumn{2}{c}{$84.6$}
&\multicolumn{2}{|c}{$83.6$} \\

\cmidrule(r){2-2}  
\cmidrule(r){3-20} 
\cmidrule(r){21-22} 

&\multicolumn{1}{|l}{{\texttt{T2I-PAL+Llama}}}
&\multicolumn{2}{|c}{$82.7$} 
&\multicolumn{2}{c}{$82.9$}
&\multicolumn{2}{c}{$84.1$}  
&\multicolumn{2}{c}{$84.6$}
&\multicolumn{2}{c}{\textbf{84.9}} 
&\multicolumn{2}{c}{$84.9$}
&\multicolumn{2}{c}{$85.4$} 
&\multicolumn{2}{c}{$85.5$}
&\multicolumn{2}{c}{\textbf{86.0}}
&\multicolumn{2}{|c}{$84.5$}  \\

&\multicolumn{1}{|l}{\cellcolor{myblue}$\textbf{{+\texttt{T2I-PAL(Ours)}}}$} 
&\multicolumn{2}{|c}{{\cellcolor{myblue}\textbf{82.7}}} 
&\multicolumn{2}{c}{{\cellcolor{myblue}\textbf{83.0}}}
&\multicolumn{2}{c}{{\cellcolor{myblue}\textbf{84.5}}}
&\multicolumn{2}{c}{{\cellcolor{myblue}\textbf{84.6}}}
&\multicolumn{2}{c}{{\cellcolor{myblue}\textbf{84.8}}}
&\multicolumn{2}{c}{{\cellcolor{myblue}\textbf{85.0}}}
&\multicolumn{2}{c}{{\cellcolor{myblue}\textbf{85.6}}}
&\multicolumn{2}{c}{{\cellcolor{myblue}\textbf{85.8}}}
&\multicolumn{2}{c}{{\cellcolor{myblue}\textbf{85.9}}}
&\multicolumn{2}{|c}{{\cellcolor{myblue}\textbf{84.7}}}
\\\hline\hline

\multirow{7}{*}{\rotatebox{90}{\textbf{VOC 2007}}}

&\multicolumn{1}{|l}{SSGRL~\cite{chen2019learning}~${\color{gray}_{\text{[ICCV}19]}}$}
&\multicolumn{2}{|c}{$77.7$} 
&\multicolumn{2}{c}{$87.6$}
&\multicolumn{2}{c}{$89.9$}  
&\multicolumn{2}{c}{$90.7$}
&\multicolumn{2}{c}{$91.4$}  
&\multicolumn{2}{c}{$91.8$}
&\multicolumn{2}{c}{$91.9$}  
&\multicolumn{2}{c}{$92.2$}
&\multicolumn{2}{c}{$92.2$}  
&\multicolumn{2}{|c}{$89.5$}  
\\

&\multicolumn{1}{|l}{GCN-ML~\cite{chen2019multi}~${\color{gray}_{\text{[CVPR}19]}}$}
&\multicolumn{2}{|c}{$74.5$} 
&\multicolumn{2}{c}{$87.4$}
&\multicolumn{2}{c}{$89.7$}  
&\multicolumn{2}{c}{$90.7$}
&\multicolumn{2}{c}{$91.0$}  
&\multicolumn{2}{c}{$91.3$}
&\multicolumn{2}{c}{$91.5$}  
&\multicolumn{2}{c}{$91.8$}
&\multicolumn{2}{c}{$92.0$}  
&\multicolumn{2}{|c}{$88.9$}  
\\

&\multicolumn{1}{|l}{Par.BCE~\cite{durand2019learning}~${\color{gray}_{\text{[CVPR}19]}}$}
&\multicolumn{2}{|c}{$80.7$} 
&\multicolumn{2}{c}{$88.4$}
&\multicolumn{2}{c}{$89.9$}  
&\multicolumn{2}{c}{$90.7$}
&\multicolumn{2}{c}{$91.2$}  
&\multicolumn{2}{c}{$91.8$}
&\multicolumn{2}{c}{$92.3$}  
&\multicolumn{2}{c}{$92.4$}
&\multicolumn{2}{c}{$92.5$}  
&\multicolumn{2}{|c}{$90.0$}  
\\

&\multicolumn{1}{|l}{SARB~\cite{pu2022semantic}~${\color{gray}_{\text{[AAAI}22]}}$}
&\multicolumn{2}{|c}{$83.5$} 
&\multicolumn{2}{c}{$88.6$}
&\multicolumn{2}{c}{$90.7$}  
&\multicolumn{2}{c}{$91.4$}
&\multicolumn{2}{c}{$91.9$}  
&\multicolumn{2}{c}{$92.2$}
&\multicolumn{2}{c}{$92.6$}  
&\multicolumn{2}{c}{$92.8$}
&\multicolumn{2}{c}{$92.9$}  
&\multicolumn{2}{|c}{$90.7$}  
\\

&\multicolumn{1}{|l}{DualCoOp~\cite{sun2022dualcoop}~${\color{gray}_{\text{[NeurIPS}22]}}$}
&\multicolumn{2}{|c}{$90.3$} 
&\multicolumn{2}{c}{$92.2$}
&\multicolumn{2}{c}{$92.8$}  
&\multicolumn{2}{c}{$93.3$}
&\multicolumn{2}{c}{$93.6$}  
&\multicolumn{2}{c}{$93.9$}
&\multicolumn{2}{c}{$94.0$}  
&\multicolumn{2}{c}{$94.1$}
&\multicolumn{2}{c}{$94.2$}  
&\multicolumn{2}{|c}{$93.2$}  
\\

&\multicolumn{1}{|l}{+TaI-DPT~\cite{guo2022texts}~${\color{gray}_{\text{[CVPR}23]}}$}
&\multicolumn{2}{|c}{$93.3$} 
&\multicolumn{2}{c}{$94.6$}
&\multicolumn{2}{c}{$94.8$}  
&\multicolumn{2}{c}{$94.9$}
&\multicolumn{2}{c}{\textbf{95.1}}  
&\multicolumn{2}{c}{$95.0$}
&\multicolumn{2}{c}{{$95.1$}}  
&\multicolumn{2}{c}{$95.3$}
&\multicolumn{2}{c}{\textbf{95.5}}
&\multicolumn{2}{|c}{$94.8$}  \\ 

&\multicolumn{1}{|l}{Positive CoOp}~\cite{rawlekar2024rethinking}~${\color{gray}_{\text{[arXiv}24]}}$
&\multicolumn{2}{|c}{$91.4$} 
&\multicolumn{2}{c}{$92.8$}
&\multicolumn{2}{c}{$93.4$}  
&\multicolumn{2}{c}{$93.6$}
&\multicolumn{2}{c}{$93.8$} 
&\multicolumn{2}{c}{$94.0$}
&\multicolumn{2}{c}{$94.2$} 
&\multicolumn{2}{c}{$94.2$}
&\multicolumn{2}{c}{$94.3$}
&\multicolumn{2}{|c}{$93.6$}  \\ 

&\multicolumn{1}{|l}{PVP}~\cite{wu2024tai}~${\color{gray}_{\text{[arXiv}24]}}$
&\multicolumn{2}{|c}{\textbf{93.7}} 
&\multicolumn{2}{c}{$94.4$}
&\multicolumn{2}{c}{$94.7$}  
&\multicolumn{2}{c}{\textbf{95.1}}
&\multicolumn{2}{c}{\textbf{95.1}} 
&\multicolumn{2}{c}{\textbf{95.2}}
&\multicolumn{2}{c}{{$95.2$}}  
&\multicolumn{2}{c}{$95.3$}
&\multicolumn{2}{c}{$95.3$}
&\multicolumn{2}{|c}{$94.9$}  \\ 

&\multicolumn{1}{|l}{Dong \textit{et al}.}~\cite{kang2024class}~${\color{gray}_{\text{[OpenReview}24]}}$
&\multicolumn{2}{|c}{$92.5$} 
&\multicolumn{2}{c}{$93.9$}
&\multicolumn{2}{c}{$94.3$}  
&\multicolumn{2}{c}{$94.7$}
&\multicolumn{2}{c}{$94.9$}  
&\multicolumn{2}{c}{$95.0$}
&\multicolumn{2}{c}{$95.1$} 
&\multicolumn{2}{c}{$95.2$}
&\multicolumn{2}{c}{$95.1$}
&\multicolumn{2}{|c}{$94.5$}  \\

\cmidrule(r){2-2}  
\cmidrule(r){3-20} 
\cmidrule(r){21-22} 

&\multicolumn{1}{|l}{{\texttt{T2I-PAL+Llama}}}
&\multicolumn{2}{|c}{$93.5$} 
&\multicolumn{2}{c}{$94.8$}
&\multicolumn{2}{c}{$94.7$}  
&\multicolumn{2}{c}{\textbf{95.0}}
&\multicolumn{2}{c}{$94.8$}  
&\multicolumn{2}{c}{$94.9$}
&\multicolumn{2}{c}{$95.4$}  
&\multicolumn{2}{c}{$95.5$}
&\multicolumn{2}{c}{$95.3$}
&\multicolumn{2}{|c}{$94.9$}  \\

&\multicolumn{1}{|l}{\cellcolor{myblue}$\textbf{{+\texttt{T2I-PAL(Ours)}}}$} 
&\multicolumn{2}{|c}{{\cellcolor{myblue}\textbf{93.7}}} 
&\multicolumn{2}{c}{{\cellcolor{myblue}\textbf{94.8}}}
&\multicolumn{2}{c}{{\cellcolor{myblue}\textbf{94.8}}}
&\multicolumn{2}{c}{{\cellcolor{myblue}\textbf{94.9}}}
&\multicolumn{2}{c}{{\cellcolor{myblue}{$94.9$}}}
&\multicolumn{2}{c}{{\cellcolor{myblue}\textbf{95.2}}}
&\multicolumn{2}{c}{{\cellcolor{myblue}\textbf{95.5}}}
&\multicolumn{2}{c}{{\cellcolor{myblue}\textbf{95.5}}}
&\multicolumn{2}{c}{{\cellcolor{myblue}\textbf{95.5}}}
&\multicolumn{2}{|c}{{\cellcolor{myblue}\textbf{95.0}}}
\\\hline\hline

\multirow{3}{*}{\rotatebox{90}{\textbf{NUS}}}

&\multicolumn{1}{|l}{DualCoOp~\cite{sun2022dualcoop}~${\color{gray}_{\text{[NeurIPS}22]}}$}
&\multicolumn{2}{|c}{$54.0$} 
&\multicolumn{2}{c}{$56.2$}
&\multicolumn{2}{c}{$56.9$}  
&\multicolumn{2}{c}{$57.4$}
&\multicolumn{2}{c}{$57.9$}  
&\multicolumn{2}{c}{$57.9$}
&\multicolumn{2}{c}{$57.6$}  
&\multicolumn{2}{c}{$58.2$}
&\multicolumn{2}{c}{$58.8$}  
&\multicolumn{2}{|c}{$57.2$}  
\\

&\multicolumn{1}{|l}{+TaI-DPT~\cite{guo2022texts}~${\color{gray}_{\text{[CVPR}23]}}$}
&\multicolumn{2}{|c}{$56.4$} 
&\multicolumn{2}{c}{$57.9$}
&\multicolumn{2}{c}{$57.8$}  
&\multicolumn{2}{c}{$58.1$}
&\multicolumn{2}{c}{$58.5$}  
&\multicolumn{2}{c}{$58.8$}
&\multicolumn{2}{c}{$58.6$}  
&\multicolumn{2}{c}{$59.1$}
&\multicolumn{2}{c}{$59.4$}
&\multicolumn{2}{|c}{$58.3$}  \\ 

&\multicolumn{1}{|l}{Dong \textit{et al}.}~\cite{kang2024class}~${\color{gray}_{\text{[OpenReview}24]}}$
&\multicolumn{2}{|c}{$55.0$} 
&\multicolumn{2}{c}{$56.9$}
&\multicolumn{2}{c}{$57.7$} 
&\multicolumn{2}{c}{$58.2$}
&\multicolumn{2}{c}{$58.6$}  
&\multicolumn{2}{c}{$58.6$}
&\multicolumn{2}{c}{$58.8$} 
&\multicolumn{2}{c}{$59.2$}
&\multicolumn{2}{c}{\textbf{59.5}}
&\multicolumn{2}{|c}{$58.1$}
\\

\cmidrule(r){2-2}  
\cmidrule(r){3-20} 
\cmidrule(r){21-22} 

&\multicolumn{1}{|l}{{\texttt{T2I-PAL+Llama}}}
&\multicolumn{2}{|c}{\textbf{56.8}}
&\multicolumn{2}{c}{\textbf{58.0}}
&\multicolumn{2}{c}{$57.8$}  
&\multicolumn{2}{c}{$58.2$}
&\multicolumn{2}{c}{$58.7$}  
&\multicolumn{2}{c}{$59.0$}
&\multicolumn{2}{c}{$59.3$}  
&\multicolumn{2}{c}{$59.3$}
&\multicolumn{2}{c}{$59.2$}
&\multicolumn{2}{|c}{$58.5$}  \\

&\multicolumn{1}{|l}{\cellcolor{myblue}$\textbf{{+\texttt{T2I-PAL(Ours)}}}$} 
&\multicolumn{2}{|c}{{\cellcolor{myblue}\textbf{56.7}}} 
&\multicolumn{2}{c}{{\cellcolor{myblue}\textbf{57.9}}}
&\multicolumn{2}{c}{{\cellcolor{myblue}\textbf{57.9}}}
&\multicolumn{2}{c}{{\cellcolor{myblue}\textbf{58.3}}}
&\multicolumn{2}{c}{{\cellcolor{myblue}\textbf{58.7}}}
&\multicolumn{2}{c}{{\cellcolor{myblue}\textbf{59.2}}}
&\multicolumn{2}{c}{{\cellcolor{myblue}\textbf{59.3}}}
&\multicolumn{2}{c}{{\cellcolor{myblue}\textbf{59.3}}}
&\multicolumn{2}{c}{{\cellcolor{myblue}{$59.3$}}}
&\multicolumn{2}{|c}{{\cellcolor{myblue}\textbf{58.5}}}
\\

\bottomrule
\end{tabular}
\end{table*}


\vspace{4pt}
\noindent\textbf{Datasets.} Our proposed method T2I-PAL is evaluated on three datasets, \ie, {VOC2007}~\cite{everingham2010pascal}, {MS-COCO}~\cite{lin2014microsoft}, and {NUS-WIDE}~\cite{chua2009nus}. As there are no training images in our method, we adopt their official \texttt{test} set to evaluate our method. For \texttt{training}, we obtain the text captions from training set of {MS-COCO}~\cite{lin2014microsoft} for both {VOC2007}~\cite{everingham2010pascal}, and {MS-COCO}~\cite{lin2014microsoft} datasets. For {NUS-WIDE}~\cite{chua2009nus}, the localized narratives from OpenImages~\cite{krasin2017openimages} are adopted to cover all the concepts in this dataset.

\vspace{4pt}
\noindent\textbf{Baselines.} To investigate the effectiveness of our method, we conduct experiments on three scenarios, \ie, \texttt{zero-shot} setting, where the models recognize new classes that have never been seen before; \texttt{few-shot} setting, where the models infer new classes from a small number of training examples and a large amount of unlabeled data, \texttt{partial-label} setting, where only some of the categories are annotated in each training image. 
%
%
For the \texttt{zero-shot} setting, we compare our method with ZSCLIP~\cite{radford2021learning}, zero-shot CLIP model for MLR; and TaI~\cite{guo2022texts}, treating text as an image in prompt tuning for MLR. 
For the \texttt{few-shot} setting, we compare our method with 
%
%
LaSO~\cite{alfassy2019laso}, a few-shot MLR method that synthesizes samples with multiple labels for generalizing to the unseen labels unseen during training; ML-FSL~\cite{simon2022meta}; CoOp~\cite{zhou2022learning}, a CLIP-based method that models a prompt's context words with learnable vectors. 
For the \texttt{partial-label} setting, we compare our method with the following baselines: two graph-based methods, \ie, 
%
SSGRL~\cite{chen2019learning}, a semantic-specific graph representation learning framework that consists of a semantic decoupling module and a semantic interaction module for MLR; GCN-ML~\cite{chen2019multi}, an MLR method that uses the directed graph to map the label graph into a set of inter-dependent object classifiers; 
Par.BCE~\cite{durand2019learning}, an MLR method that uses the BCE loss and instance-prototype-level semantic representations to exploit the proportion of known labels per image; SARB~\cite{pu2022semantic}, a unified semantic-aware representation blending framework, leverages both instance-level and prototype-level semantic representations to enhance the labeling of unknown instances; 
DualCoOp~\cite{sun2022dualcoop}, a MLR method that utilizes the strong pre-trained CLIP and dual context optimization mechanism as a unified framework for both partial-label and zero-shot MLR; +TaI-DPT~\cite{guo2022texts}, an ensemble method that integrates TaI~\cite{guo2022texts} with the partial-label MLR method, DualCoOp~\cite{sun2022dualcoop};
%
Positive CoOp~\cite{rawlekar2024rethinking}, a method in which only one prompt is learned to compare the positive and negative prompts in a Vision-Language Model (VLM) for MLR. 
PVP~\cite{wu2024tai}, an MLR method that employs pseudo-visual prompts to minimize the diverse visual knowledge in the pre-trained VLM. 
Dong \textit{et al}.~\cite{kang2024class}, an MLR method that utilizes class concept representations to learn rich contexts from the extensive descriptions of images.

\begin{table}
\renewcommand{\arraystretch}{1.3}
	\caption{\small \textbf{Comparison} against various SOTAs under \texttt{few-shot} setting on MS-COCO dataset with $16$ \textbf{novel} classes.}
	\label{tab3}
        \setlength{\tabcolsep}{8.3pt}
	\fontsize{9}{9}\selectfont
	\centering
	\begin{tabular}{l  c c c}
\toprule
{Method}
&\multicolumn{1}{c}{\textbf{0-Shot}}
&\multicolumn{1}{c}{\textbf{1-Shot}} 
&\multicolumn{1}{c}{\textbf{5-Shot}}\\ 

\cmidrule(r){1-1} \cmidrule(lr){2-2} \cmidrule(lr){3-3} \cmidrule(lr){4-4} 

LaSO~\cite{alfassy2019laso} ${\color{gray}_{\text{[CVPR}19]}}$
&\multicolumn{1}{|c}{$-$} 
&\multicolumn{1}{c}{$45.3$}
&\multicolumn{1}{c}{$58.1$}  \\

ML-FSL~\cite{simon2022meta}~${\color{gray}_{\text{[WACV}22]}}$
&\multicolumn{1}{|c}{$-$} 
&\multicolumn{1}{c}{$54.4$}
&\multicolumn{1}{c}{$63.6$}  \\

TaI-DPT~\cite{guo2022texts}~${\color{gray}_{\text{[CVPR}23]}}$
&\multicolumn{1}{|c}{$59.2$} 
&\multicolumn{1}{c}{$-$}
&\multicolumn{1}{c}{$-$}  \\

\multicolumn{1}{l}{Dong \textit{et al}.}~\cite{kang2024class}~${\color{gray}_{\text{[arXiv}24]}}$
&\multicolumn{1}{|c}{$61.4$}
&\multicolumn{1}{c}{$-$}
&\multicolumn{1}{c}{$-$}  \\

\multicolumn{1}{l}{PVP}~\cite{wu2024tai}~${\color{gray}_{\text{[arXiv}24]}}$
&\multicolumn{1}{|c}{$64.4$}
&\multicolumn{1}{c}{$-$}
&\multicolumn{1}{c}{$-$} \\

\cmidrule(r){1-1} \cmidrule(lr){2-4}

\multicolumn{1}{l}{{\texttt{T2I-PAL+Llama}}}
&\multicolumn{1}{|c}{$66.2$}
&\multicolumn{1}{c}{$-$}
&\multicolumn{1}{c}{$-$}  \\

{\cellcolor{myblue}$\textbf{{\texttt{T2I-PAL(Ours)}}}$} 
&\multicolumn{1}{|c}{{\cellcolor{myblue}\textbf{66.3}\stdvu{$\underline{{1.9}}$}}} 
&\multicolumn{1}{c}{{\cellcolor{myblue}$-$}}
&\multicolumn{1}{c}{{\cellcolor{myblue}$-$}}\\

\bottomrule
\end{tabular}
\end{table}

\begin{table*}
\renewcommand{\arraystretch}{1.3} 
\setlength{\tabcolsep}{13pt} 
	\caption{\small \textbf{Comparison} with state-of-the-arts under the \texttt{few-shot} setting, where +TaI-DPT~\cite{guo2022texts} and +T2I-PAL indicate integration with the few-shot MLR method, CoOp~\cite{zhou2022learning}, respectively. ${\color{ForestGreen}\uparrow}$ indicates \textbf{improvements} compared with the \textbf{top-1} ranked baseline method, \ie, +TaI-DPT~\cite{guo2022texts}.}
	\label{tab4}
	\fontsize{9}{9}\selectfont
	\centering
	\begin{tabular}{c l  c c c c c c}
\toprule
&Method
&\multicolumn{1}{c}{\textbf{0-Shot}}
&\multicolumn{1}{c}{\textbf{1-Shot}} 
&\multicolumn{1}{c}{\textbf{2-Shot}}
&\multicolumn{1}{c}{\textbf{4-Shot}}
&\multicolumn{1}{c}{\textbf{8-Shot}}
&\multicolumn{1}{c}{\textbf{16-Shot}}
\\ 

\cmidrule(r){2-2} \cmidrule(lr){3-3} \cmidrule(lr){4-4}\cmidrule(lr){5-5}\cmidrule(lr){6-6}\cmidrule(lr){7-7}\cmidrule(lr){8-8}

\multirow{6}{*}{\rotatebox{90}{MS-COCO} }
&\multicolumn{1}{|l}{ZSCLIP~\cite{radford2021learning}~${\color{gray}_{\text{[ICML}21]}}$}
&\multicolumn{1}{|c}{$47.3$} 
&\multicolumn{1}{c}{$-$}
&\multicolumn{1}{c}{$-$} 
&\multicolumn{1}{c}{$-$} 
&\multicolumn{1}{c}{$-$}
&\multicolumn{1}{c}{$-$}
\\

&\multicolumn{1}{|l}{CoOp~\cite{zhou2022learning}~${\color{gray}_{\text{[IJCV}22]}}$}
&\multicolumn{1}{|c}{$-$} 
&\multicolumn{1}{c}{$52.6$}
&\multicolumn{1}{c}{$57.3$}
&\multicolumn{1}{c}{$58.1$} 
&\multicolumn{1}{c}{$59.2$}
&\multicolumn{1}{c}{$59.8$}
\\

&\multicolumn{1}{|l}{TaI-DPT~\cite{guo2022texts}~${\color{gray}_{\text{[CVPR}23]}}$}
&\multicolumn{1}{|c}{$65.1$} 
&\multicolumn{1}{c}{$-$}
&\multicolumn{1}{c}{$-$} 
&\multicolumn{1}{c}{$-$} 
&\multicolumn{1}{c}{$-$}
&\multicolumn{1}{c}{$-$}
\\

&\multicolumn{1}{|l}{+TaI-DPT~\cite{guo2022texts}~${\color{gray}_{\text{[CVPR}23]}}$}
&\multicolumn{1}{|c}{$-$} 
&\multicolumn{1}{c}{$65.8$}
&\multicolumn{1}{c}{$66.2$} 
&\multicolumn{1}{c}{$67.6$} 
&\multicolumn{1}{c}{$68.1$}
&\multicolumn{1}{c}{$68.9$}
\\

\cmidrule(r){2-2} \cmidrule(lr){3-8} 

&\multicolumn{1}{|l}{{\texttt{T2I-PAL+Llama}}}
&\multicolumn{1}{|c}{$71.4$}
&\multicolumn{1}{c}{$-$}
&\multicolumn{1}{c}{\textbf{}$-$}
&\multicolumn{1}{c}{$-$}
&\multicolumn{1}{c}{$-$}
&\multicolumn{1}{c}{$-$}
\\

&\multicolumn{1}{|l}{{\texttt{+T2I-PAL+Llama}}}   
&\multicolumn{1}{|c}{$-$} 
&\multicolumn{1}{c}{$71.5$}
&\multicolumn{1}{c}{$71.8$}
&\multicolumn{1}{c}{$73.0$}
&\multicolumn{1}{c}{$73.5$}
&\multicolumn{1}{c}{$74.1$}
\\

&\multicolumn{1}{|l}{\cellcolor{myblue}$\textbf{{\texttt{T2I-PAL(Ours)}}}$} 
&\multicolumn{1}{|c}{{\cellcolor{myblue}\textbf{71.4}\stdvu{$\underline{{6.3}}$}}} 
&\multicolumn{1}{c}{{\cellcolor{myblue}$-$}}
&\multicolumn{1}{c}{{\cellcolor{myblue}$-$}}
&\multicolumn{1}{c}{{\cellcolor{myblue}$-$}}
&\multicolumn{1}{c}{{\cellcolor{myblue}$-$}}
&\multicolumn{1}{c}{{\cellcolor{myblue}$-$}}\\ 

&\multicolumn{1}{|l}{\cellcolor{myblue}+$\textbf{{\texttt{T2I-PAL(Ours)}}}$} 
&\multicolumn{1}{|c}{{\cellcolor{myblue}$-$}}
&\multicolumn{1}{c}{{\cellcolor{myblue}\textbf{71.6}\stdvu{$\underline{{5.8}}$}}}
&\multicolumn{1}{c}{{\cellcolor{myblue}\textbf{71.8}\stdvu{$\underline{{5.6}}$}}}
&\multicolumn{1}{c}{{\cellcolor{myblue}\textbf{73.1}\stdvu{$\underline{{5.5}}$}}} 
&\multicolumn{1}{c}{{\cellcolor{myblue}\textbf{73.5}\stdvu{$\underline{{5.4}}$}}} 
&\multicolumn{1}{c}{{\cellcolor{myblue}\textbf{74.1}\stdvu{$\underline{{5.2}}$}}}\\ \hline\hline


\multirow{6}{*}{\rotatebox{90}{{VOC-2007}}} 
&\multicolumn{1}{|l}{ZSCLIP~\cite{radford2021learning}~${\color{gray}_{\text{[ICML}21]}}$}
&\multicolumn{1}{|c}{$76.2$} 
&\multicolumn{1}{c}{$-$}
&\multicolumn{1}{c}{$-$} 
&\multicolumn{1}{c}{$-$} 
&\multicolumn{1}{c}{$-$}
&\multicolumn{1}{c}{$-$}
\\

&\multicolumn{1}{|l}{CoOp\cite{zhou2022learning}~${\color{gray}_{\text{[IJCV}22]}}$}
&\multicolumn{1}{|c}{$-$} 
&\multicolumn{1}{c}{$79.3$}
&\multicolumn{1}{c}{$83.2$}
&\multicolumn{1}{c}{$83.8$} 
&\multicolumn{1}{c}{$84.5$}
&\multicolumn{1}{c}{$85.7$}
\\

&\multicolumn{1}{|l}{TaI-DPT~\cite{guo2022texts}~${\color{gray}_{\text{[CVPR}23]}}$}
&\multicolumn{1}{|c}{$88.3$} 
&\multicolumn{1}{c}{$-$}
&\multicolumn{1}{c}{$-$} 
&\multicolumn{1}{c}{$-$} 
&\multicolumn{1}{c}{$-$}
&\multicolumn{1}{c}{$-$}
\\

&\multicolumn{1}{|l}{+TaI-DPT~\cite{guo2022texts}~${\color{gray}_{\text{[CVPR}23]}}$}
&\multicolumn{1}{|c}{$-$} 
&\multicolumn{1}{c}{$88.6$}
&\multicolumn{1}{c}{$89.2$} 
&\multicolumn{1}{c}{$89.1$} 
&\multicolumn{1}{c}{$89.5$}
&\multicolumn{1}{c}{$90.1$}
\\

\cmidrule(r){2-2} \cmidrule(lr){3-8} 

&\multicolumn{1}{|l}{{\texttt{T2I-PAL+Llama}}}
&\multicolumn{1}{|c}{\textbf{91.6}}
&\multicolumn{1}{c}{$-$}
&\multicolumn{1}{c}{\textbf{}$-$}
&\multicolumn{1}{c}{$-$}
&\multicolumn{1}{c}{$-$}
&\multicolumn{1}{c}{$-$}
\\

&\multicolumn{1}{|l}{{\texttt{+T2I-PAL+Llama}}}  
&\multicolumn{1}{|c}{$-$} 
&\multicolumn{1}{c}{$91.7$}
&\multicolumn{1}{c}{$92.0$}
&\multicolumn{1}{c}{$92.2$} 
&\multicolumn{1}{c}{$92.2$}
&\multicolumn{1}{c}{$92.9$}
\\

&\multicolumn{1}{|l}{\cellcolor{myblue}$\textbf{{\texttt{T2I-PAL(Ours)}}}$} 
&\multicolumn{1}{|c}{{\cellcolor{myblue}\textbf{91.5}\stdvu{$\underline{{3.2}}$}}} 
&\multicolumn{1}{c}{{\cellcolor{myblue}$-$}}
&\multicolumn{1}{c}{{\cellcolor{myblue}$-$}}
&\multicolumn{1}{c}{{\cellcolor{myblue}$-$}}
&\multicolumn{1}{c}{{\cellcolor{myblue}$-$}}
&\multicolumn{1}{c}{{\cellcolor{myblue}$-$}}\\

&\multicolumn{1}{|l}{\cellcolor{myblue}+$\textbf{{\texttt{T2I-PAL(Ours)}}}$} 
&\multicolumn{1}{|c}{{\cellcolor{myblue}$-$}}
&\multicolumn{1}{c}{{\cellcolor{myblue}\textbf{91.7}\stdvu{$\underline{{3.1}}$}}}
&\multicolumn{1}{c}{{\cellcolor{myblue}\textbf{92.1}\stdvu{$\underline{{2.9}}$}}}
&\multicolumn{1}{c}{{\cellcolor{myblue}\textbf{92.2}\stdvu{$\underline{{3.1}}$}}} 
&\multicolumn{1}{c}{{\cellcolor{myblue}\textbf{92.3}\stdvu{$\underline{{2.8}}$}}} 
&\multicolumn{1}{c}{{\cellcolor{myblue}\textbf{92.9}\stdvu{$\underline{{2.8}}$}}}\\ 

\bottomrule
\end{tabular}
\end{table*}

\subsection{Comparison with State-of-the-Arts}\label{sota}
Table~\ref{tab1} and Table~\ref{tab2} summarize the mAP values of the \texttt{zero-shot} and \texttt{partial label} settings over three datasets, \ie, VOC 2007~\cite{everingham2010pascal}, MS-COCO~\cite{lin2014microsoft}, and NUS-WIDE~\cite{chua2009nus}, where 
+TaI-DPT indicates integrating TaI-DPT~\cite{guo2022texts} with the partial-label MLR method, DualCoOp~\cite{sun2022dualcoop}.
As can be seen from these tables, the baseline performance of the partial labeled in Table~\ref{tab2} is generally better than the zero-shot baseline methods in Table~\ref{tab1}, mainly because the labeled training data can make the model perform better on the test set. 
Under the \texttt{zero-shot} setting, our method outperforms the top-ranked TaI-DPT~\cite{guo2022texts} by $0.6$\%, $1.5$\%, and $0.8$\% on the three datasets, respectively. 
%
%
More importantly, even when enhanced with the partial-label MLR method, DualCoOp~\cite{sun2022dualcoop}, the strongest baseline, Dong \textit{et al.}~\cite{kang2024class}, remains inferior to our method. For example, with $70$\% label annotation, our method achieves $\textbf{85.6}$ \textit{vs.} $84.4$ on MS-COCO~\cite{lin2014microsoft}, $\textbf{95.5}$ \textit{vs.} $95.1$ on VOC 2007~\cite{everingham2010pascal}, and $\textbf{59.3}$ \textit{vs.} $58.8$ on NUS-WIDE~\cite{chua2009nus}.
%
%
These results support our conclusion that leveraging pre-trained text-to-image generation models to generate photo-realistic and diverse images from text captions is beneficial for reducing the modality gap.
We further evaluate the effectiveness of our T2I-PAL with the various state-of-the-arts in the \texttt{few-shot} setting. Following existing few-shot MLR methods~\cite{alfassy2019laso,simon2022meta}, a model was trained on known classes while deployed to $16$ novel classes. In Table~\ref{tab3}, we record various few-shot MLR methods with zero-shot TaI-DP~\cite{guo2022texts} and T2I-PAL on the $16$ novel classes. As can be seen from this table, the performance of T2I-PAL is higher than that of TaI-DPT, even \textbf{surpassing} the top-$1$ rank method trained on $5$ shot samples, \ie, ML-FSL/$5$-shot~\cite{simon2022meta}: $63.6$ \textit{vs.} T2I-PAL/$0$-shot: $\textbf{66.3}$. 

Additionally, following~\cite{guo2022texts}, we adopt the strategy in~\cite{alfassy2019laso} to treat all classes as novel classes and select $1$, $2$, $4$, $8$, and $16$ shot samples for each class. Since neither TaI-DPT~\cite{guo2022texts} nor our method, T2I-PAL, has original annotated images, we integrate them with CoOp to record the performance under different \texttt{few-shot} settings, termed +TaI-DPT~\cite{guo2022texts} and +T2I-PAL. 
As can be seen in Table~\ref{tab4}, although the integration of TaI-DPT~\cite{guo2022texts} and CoOp~\cite{zhou2022learning}, +TaI-DPT~\cite{guo2022texts}, improves the average performance of CoOp~\cite{zhou2022learning} in the few-shot of the two datasets, \ie, $57.4$ $\rightarrow$ $\textbf{67.3}$ on MS-COCO~\cite{lin2014microsoft}, and $83.3$ $\rightarrow$ $\textbf{89.3}$ on VOC 2007~\cite{everingham2010pascal}, our method still gets further improvement, \ie, $67.3$ $ \rightarrow$ $\textbf{72.8}$ on MS-COCO~\cite{lin2014microsoft}, and $89.3$ $\rightarrow$ $\textbf{92.2}$ on VOC 2007~\cite{everingham2010pascal}. 
More importantly, the performance of our method on zero-shot has surpassed the performance on $16$-shot after the integration of TaI-DPT and CoOp~\cite{zhou2022learning}, \eg, $68.9$ \textit{vs.} $\textbf{71.4} $ on MS-COCO~\cite{lin2014microsoft}, and $90.1$ \textit{vs.} $\textbf{91.5}$ on VOC 2007~\cite{everingham2010pascal}. 
The outstanding performance of T2I-PAL in the few-shot setting again confirms our core idea that learning prompts and adapters using synthesized images through text captions can help tackle the modality gap issue when using only text captions for PEFT.
Besides, we also included the performance of T2I-PAL+LLaMA in those tables. From these tables, it is evident that the captions generated by LLaMA continue to demonstrate good performance. Since the captions generated by LLaMA also include the corresponding categories and maintain the same quantity as the original captions, the amount of information provided is consistent.

\begin{figure}
	\begin{center}
		\includegraphics[width=\linewidth]{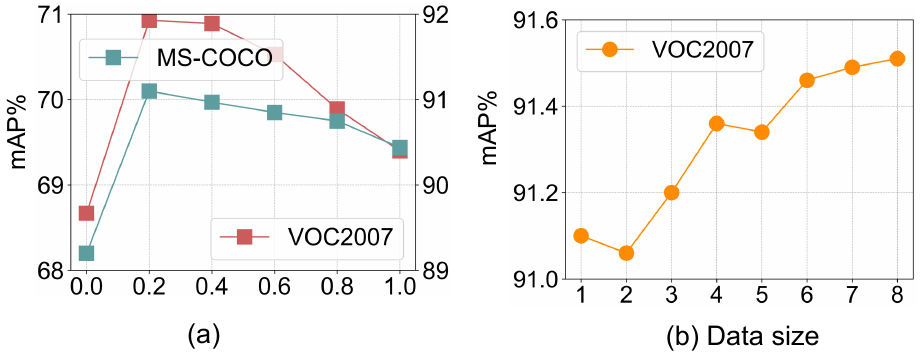}
         \put(-180,4){ \small{$\gamma$}}
	\end{center}
	\captionsetup{font=small}
	\caption{\small\textbf{(a)} \textbf{Analysis} with regard to the different values of $\gamma$, where the \textit{larger} the value of the $\gamma$, the larger the proportion of the \textit{synthetic image} in our method, vice versa; \textbf{(d)} \textbf{Ablations} on the different \textbf{size of synthetic} data on our method.}
	\label{fig3new}
\end{figure}

\subsection{Effectiveness on the Different Tasks}
To evaluate our proposed method, we conducted experiments on zero-shot image classification tasks under two scenarios: single-label classification and multi-label attribute classification~\cite{kao2024enhancingclipconceptualembedding}. As shown in Table \ref{danbiaoqian}, both TaI-DPT and our T2I-PAL method significantly outperform CLIP, highlighting the advantage of leveraging caption information. Notably, T2I-PAL use of multimodal data (synthetic images and text captions) further improves performance in both single-label and multi-label attribute classification.

\begin{table}[t]
\renewcommand{\arraystretch}{1.3}
	\caption{\small \textbf{Comparison} with state-of-the-art methods for single-label classification (Flowers102, Food101) and multi-label attribute classification (AWA2).}
	\label{danbiaoqian}
        \setlength{\tabcolsep}{11.6pt}
	\fontsize{9}{9}\selectfont
	\centering
	\begin{tabular}{ l  cc cc cc}
\toprule
{Method}
&\multicolumn{2}{c}{ZSCLIP}
&\multicolumn{2}{c}{{TaI-DPT}} 
&\multicolumn{2}{c}{T2I-PAL}\\ 

\cmidrule(r){1-1} \cmidrule(r){2-3}  \cmidrule(r){4-5}  \cmidrule(r){6-7}  

{Flowers102 }
&\multicolumn{2}{|c}{$61.75$}
&\multicolumn{2}{c}{$83.5$}
&\multicolumn{2}{c}{$90.6$}
\\

{Food101  }
&\multicolumn{2}{|c}{$73.97$}
&\multicolumn{2}{c}{$86.7$}
&\multicolumn{2}{c}{$91.3$}
\\ 
\cmidrule(r){1-1} \cmidrule(r){2-3}  \cmidrule(r){4-5}  \cmidrule(r){6-7}  
{AWA2}
&\multicolumn{2}{|c}{$55.80$}
&\multicolumn{2}{c}{$63.6$}
&\multicolumn{2}{c}{$70.4$}
\\

\bottomrule
\end{tabular}
\end{table}

\subsection{Ablation Study on the Modality Gap}\label{ab1}
\vspace{2pt}

Here, we compute the cosine similarity between the text prompt \& CLS (Text.pro+CLS) and the extracted features under various conditions using the MSCOCO dataset. This evaluation assesses how synthetic images and the shared adapter help bridge the modality gap. The considered conditions include test images (Test.img), captions of the original training images (Train.cap), and synthetic images (Train.syc) after applying the adapter.
Higher values in the table indicate greater similarity and a smaller modality gap. As shown in Table~\ref{tabgap}, the modality gap between Test.img and Text.pro + CLS is the largest, as images and text belong to entirely different domains. However, replacing images with their corresponding training captions (\ie, TaI) reduces this domain gap.
Notably, our method significantly improves domain alignment even without the adapter—for example, the similarity between Train.syc and Text.pro+CLS increases from $0.3178$ to $0.3890$. More importantly, after applying the adapter, the domain gap is further reduced, with the similarity between synthetic training images and Text.pro + CLS improving from $0.3178$ to $0.6110$ compared to TaI. These results clearly demonstrate that synthetic images and the shared adapter effectively mitigate the modality gap.


\begin{table}
\renewcommand{\arraystretch}{1.3}

 \caption{\small \textbf{Modality gap} analysis, where higher values in the table indicate greater similarity and a smaller modality gap.}
	\label{tabgap}
        \setlength{\tabcolsep}{8pt}
	\fontsize{9}{9}\selectfont
	\centering

	\begin{tabular}{l  c }
\toprule
 
{Method}

&\multicolumn{1}{c}{\textbf{Cos Sim}}\\ 

\cmidrule(r){1-1} \cmidrule(lr){2-2}

Test.img \textit{vs}. Text.pro+CLS. 
&\multicolumn{1}{|c}{$0.2001$} 
 \\

Train.cap \textit{vs}. Text.pro+CLS. (TaI)~\cite{guo2022texts}~
&\multicolumn{1}{|c}{$0.3178$} 
 \\
\cmidrule(r){1-1} \cmidrule(lr){2-2}

Train.syc \textit{vs}. Text.pro+CLS. (T2I w.o/Adapter)
&\multicolumn{1}{|c}{$0.3890$} 
 \\

Train.syc \textit{vs}. Text.pro+CLS. (T2I-PAL)
&\multicolumn{1}{|c}{{\cellcolor{myblue}\textbf{0.6110}}} 
\\

\bottomrule
\end{tabular}

\end{table}

\subsection{Ablation Study on the Importance of Different Data Modalities}\label{ab1}
\vspace{2pt}
\noindent\textbf{Text \textit{vs.} Synthetic \textit{vs.} Original Images in Training Data.} Here, we investigate how the different types of {training} data influence the results. As such, we construct some variations in Table~\ref{tab5}, where \texttt{S.I}, \texttt{Text}, and \texttt{O.I} indicate that the method uses the synthesized image, text caption, and original image as the {training} data, respectively. As can be seen from this table, the performance on the \texttt{S.I} is significantly better than \texttt{Text}, \ie, TaI-DPT: $88.3$ \textit{vs.} \texttt{Ours}: (\texttt{S.I}): \textbf{90.8}. When our method employs two types of {training} data, the performance will be further improved, \ie, \texttt{Ours} (\texttt{S.I}): $90.8$ $\rightarrow$ \texttt{Ours} (Full): \textbf{91.5}. 
Additionally, \texttt{Ours}: (\texttt{Three}) can be as the upper bound (\texttt{UB.}) of our method which leverages text captions and synthetic images as well as original annotated images in training.
Nonetheless, the number of synthetic images is $6$$\times$ of the original images.
Thus, further including original images in training (\ie, \texttt{Ours}: (\texttt{Three}), $91.6$) only brings moderate gain to \texttt{Ours} (Full) ($91.5$). 
The results indicate that original images add minor diversity when the number of synthetic images is larger (\ie, $6$$\times$).
Further, we also investigate how the \textit{quality of text captions}, the \textit{synthetic images}, and \textit{additional text data} influence our method in the following.

\begin{table*}[t]
\renewcommand{\arraystretch}{1.3}
	\caption{\small\textbf{Ablation studies} with regard to the different \texttt{Adapters} of T2I-PAL on the three datasets, where ${\color{red}\downarrow}$ indicates \textbf{decrements} compared with our full model, \emph{w.} \texttt{Loc.Adp} (\texttt{Ours}).}
	\label{tab6}
        \setlength{\tabcolsep}{11.5pt}
	\fontsize{9}{9}\selectfont
	\centering
	\begin{tabular}{l  c c cc cc cc c}
\toprule

\textbf{Variation}
&\multicolumn{1}{c}{\textbf{\texttt{Glo.Adp}}}
&\multicolumn{1}{c}{\textbf{\texttt{Loc.Adp}}}
&\multicolumn{2}{l}{\textbf{~MS-COCO~}} 
&\multicolumn{2}{l}{\textbf{~VOC-2007~}}
&\multicolumn{2}{l}{\textbf{~NUS-WIDE~}} 
&\multicolumn{1}{l}{\textbf{~Average~}} 
\\
 \cmidrule(lr){1-1} \cmidrule(l){2-2} \cmidrule(l){3-3} \cmidrule(l){4-5} \cmidrule(l){6-7} \cmidrule(l){8-9} \cmidrule(l){10-10}
 
$\emph{w/o.}$ \texttt{Adp} 
&\multicolumn{1}{|c}{$-$}
&$-$
&\multicolumn{2}{|l}{$70.6$\stdvd{$\underline{0.8}$}}
&\multicolumn{2}{l}{$91.0$\stdvd{$\underline{0.5}$}}
&\multicolumn{2}{l}{$47.1$\stdvd{$\underline{0.3}$}}
&\multicolumn{1}{|l}{$69.57$\stdvd{$\underline{0.53}$}}
\\

$\emph{w.}$ \texttt{Glo.Adp}
&\multicolumn{1}{|c}{\Checkmark}
&\multicolumn{1}{c}{\Checkmark}
&\multicolumn{2}{|l}{$70.7$\stdvd{$\underline{0.7}$}}
&\multicolumn{2}{l}{$91.1$\stdvd{$\underline{0.4}$}}
&\multicolumn{2}{l}{$47.1$\stdvd{$\underline{0.3}$}}
&\multicolumn{1}{|l}{$69.63$\stdvd{$\underline{0.47}$}}\\

{\cellcolor{myblue}$\emph{w.}$ \texttt{Loc.Adp} (\texttt{Ours}) }
&\multicolumn{1}{|c}{{\cellcolor{myblue}$-$}}
&\multicolumn{1}{c}{{\cellcolor{myblue}\Checkmark}}
&\multicolumn{2}{|l}{{\cellcolor{myblue}\textbf{$71.4$}}} 
&\multicolumn{2}{l}{{\cellcolor{myblue}\textbf{$91.5$}}}
&\multicolumn{2}{l}{{\cellcolor{myblue}\textbf{$47.4$}}}
&\multicolumn{1}{|l}{{\cellcolor{myblue}\textbf{$70.1$}}}\\
\bottomrule
\end{tabular}
\end{table*}

\begin{table}
\renewcommand{\arraystretch}{1.3}
	\caption{\small\textbf{Ablation studies} with regard to the different types of \texttt{training} data, where (\texttt{UB.}) is the \textit{upper-bound} performance by using text captions, synthetic images, and original images.} 
	\label{tab5}
        \setlength{\tabcolsep}{7pt}
	\fontsize{9}{9}\selectfont
	\centering
	\begin{tabular}{l  c c c c}
\toprule

\textbf{Variation}
&\multicolumn{1}{c}{\textbf{\texttt{S.I}}}
&\multicolumn{1}{c}{\textbf{\texttt{Text}}}
&\multicolumn{1}{c}{\textbf{\texttt{O.I}}}
&\multicolumn{1}{l}{\textbf{\textbf{VOC 2007}}}
\\
\cmidrule(lr){1-1} \cmidrule(l){2-4} \cmidrule(l){5-5}

TaI-DPT
&\multicolumn{1}{|c}{$-$}
&\Checkmark
&$-$
&\multicolumn{1}{|l}{$88.3$\stdvd{$3.3$}}
\\



 
{\cellcolor{myblue}\texttt{Ours} (\texttt{S.I}) }
&\multicolumn{1}{|c}{\cellcolor{myblue}{\Checkmark}}
&\multicolumn{1}{c}{{\cellcolor{myblue}$-$}}
&{\cellcolor{myblue}$-$}
&\multicolumn{1}{|l}{{\cellcolor{myblue}$90.8$\stdvd{$\underline{0.8}$}}}
\\

{\cellcolor{myblue}\texttt{Ours} (\texttt{Full}) }
&\multicolumn{1}{|c}{{\cellcolor{myblue}\Checkmark}}
&\multicolumn{1}{c}{{\cellcolor{myblue}\Checkmark}}
&\multicolumn{1}{c}{{\cellcolor{myblue}$-$}}
&\multicolumn{1}{|l}{{\cellcolor{myblue}\textbf{$91.5$}\stdvd{${\underline{0.1}}$}}}\\
\cmidrule(lr){1-1} \cmidrule(l){2-4} \cmidrule(l){5-5}

{\cellcolor{myblue}\texttt{Ours} (\texttt{Three})(\texttt{UB}.)}
&\multicolumn{1}{|c}{{\cellcolor{myblue}\Checkmark}}
&\multicolumn{1}{c}{{\cellcolor{myblue}\Checkmark}}
&\multicolumn{1}{c}{{\cellcolor{myblue}\Checkmark}}
&\multicolumn{1}{|l}{{\cellcolor{myblue}\textbf{$91.6$}}}\\

\bottomrule
\end{tabular}

\end{table}

\vspace{4pt}
\noindent\textbf{Balance of Synthetic Images \textit{vs.} Text.} 
Since our method absorbs the complementary advantages of text captions and synthesized images from the pre-trained text-to-image model, it is necessary to investigate how these two mechanisms can help train classifiers. To this end, we performed an analysis of the effect of the hyperparameter $\gamma$ on the two datasets. The results are shown in Fig.~\ref{fig3new} (a), where the larger the value of the $\gamma$, the larger the proportion of the synthesized image, and vice versa. We can see from the figure that when $\gamma$ = $0$, that is, there is no synthesized image component in our model, the performance is lowest. Such results indicate that the synthetic image is very important for PEFT without the original annotated training image. 
Training prompts on text captions for PEFT will be greatly affected by the modality gap because training prompts in the text domain are difficult to directly adapt to the image domain. 
%
However, the performance of T2I-PAL increases rapidly with increasing $\gamma$, which is mainly the gain brought by synthetic images, \ie, directly making up the modality gap when using only text captions for PEFT. 
For example, when $\gamma$ = $0.2$, our method achieves a performance of $91.1$ and $70.9$ on the two datasets. When the $\gamma$ values continue to increase, \eg, $\gamma$ = $1$, the performance of T2I-PAL decreases to $90.4$ and $69.4$, mainly because the advantages of text caption are lost while completely adopting the synthetic images. 
Nevertheless, the performance of our method when $\gamma$ = $1$ is still higher than that of $\gamma$ = $0$, which means that the benefits of synthetic images for PEFT are much greater than those of text captions. 
Furthermore, we observe that different values of $\gamma$ perform similarly on the two datasets, which proves that T2I-PAL are robust to different datasets. To this end, we encourage adopting text captions and synthetic images to tackle the modality gap under such a scenario.
Additionally, we also explore the effect with \textit{additional text data} as well as the \textit{quality of text captions} in Section~\ref{ab2}.
To this end, we encourage adopting text captions and also leveraging pre-trained text-to-image generation models to generate photo-realistic and diverse images from text captions to tackle the modality gap under such a scenario.




\vspace{4pt}
\noindent\textbf{Potential of Synthetic Dataset Size.} Considering that synthetic data is freely available, photo-realistic, diverse, and not limited by annotations, we examine whether more synthetic visual data can help improve model performance. To this end, we record the testing results on {VOC 2007} with synthetic visual data of different sizes in Fig.~\ref{fig3new} (b). It can be seen from the figure that as the number of synthetic images increases, the performance of our method, T2I-PAL gradually improves until it stabilizes when the size reaches $6$. Given this, we set the size of the synthetic image to $6$ in our experiments in Section~\ref{sota}. It is worth noting that even when the size of the synthetic image equals $1$, the performance of the model is still much higher than the top-$1$ ranked baseline method, TaI-DPT, \ie, TaI-DPT~\cite{guo2022texts}: $88.3$ \textit{vs.} \texttt{Ours}: $\textbf{91.1}$, gains $3.2$\% improvements on {VOC 2007}. Consequently, the result indicates that our method provides an effective solution when using only text captions for MLR.
Additionally, we explore the impact of the quality of the synthesized images and the modality gap between the two images on our method in Section~\ref{ab2}.

\begin{figure}
	\begin{center}
		\includegraphics[width=\linewidth]{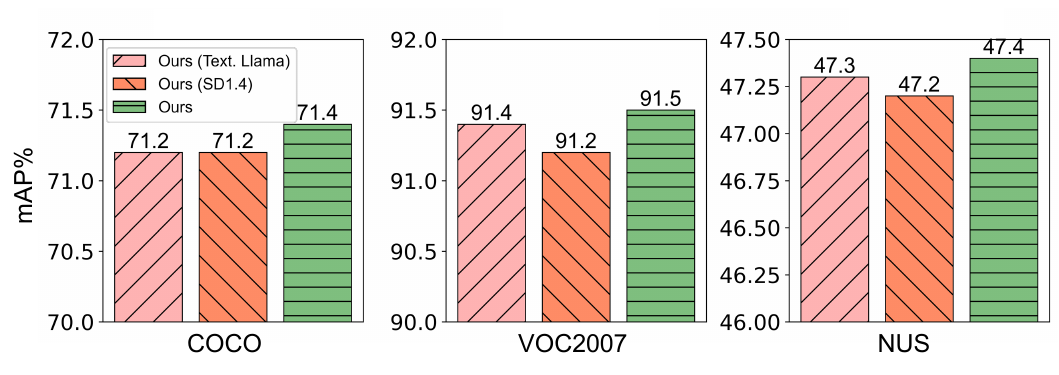}
	\end{center}
	\captionsetup{font=small}
	\caption{\small \textbf{Analysis} of the quality of \textit{text captions} and \textit{synthetic images} influence our method on the three datasets.}
	\label{cvprtext}
\end{figure}

\begin{table}\footnotesize
\renewcommand{\arraystretch}{1.3}
	\caption{\small\textbf{Ablation studies} of elimination of the modality gap.}
	\label{Amodality}
        \setlength{\tabcolsep}{6.5pt}
	\fontsize{9}{9}\selectfont
	\centering
	\begin{tabular}{l  c c c c }
\toprule
\textbf{Methods}
&\multicolumn{1}{c}{{MS-COCO}}
&\multicolumn{1}{l}{VOC 2007}
&\multicolumn{1}{l}{{NUS-WIDE}}
\\
\cmidrule(lr){1-1}
\cmidrule(l){2-4} 

{\texttt{Ours}}(1$\times${\texttt{S.I}})
&\multicolumn{1}{c}{\textbf{$70.5$}}
&\multicolumn{1}{c}{\textbf{$90.8$}}
&\multicolumn{1}{c}{\textbf{$46.8$}}\\

{\texttt{Ours}}({\texttt{O.I-40\%}})
&\multicolumn{1}{c}{\textbf{$70.5$}}
&\multicolumn{1}{c}{\textbf{$90.9$}}
&\multicolumn{1}{c}{\textbf{$46.8$}}
\\

{\texttt{Ours}}({\texttt{O.I-80\%}})
&\multicolumn{1}{c}{\textbf{$70.6$}}
&\multicolumn{1}{c}{\textbf{$90.8$}}
&\multicolumn{1}{c}{\textbf{$46.9$}}\\

{\texttt{Ours}}({\texttt{O.I}})
&\multicolumn{1}{c}{\textbf{$70.7$}}
&\multicolumn{1}{c}{\textbf{$90.9$}}
&\multicolumn{1}{c}{\textbf{$46.9$}}\\

{\cellcolor{myblue}{\texttt{Ours}}(6$\times${\texttt{S.I}})}
&\multicolumn{1}{c}{{\cellcolor{myblue}\textbf{$71.4$}}}
&\multicolumn{1}{c}{{\cellcolor{myblue}\textbf{$91.5$}}}
&\multicolumn{1}{c}{{\cellcolor{myblue}\textbf{$47.4$}}}\\

\bottomrule
\end{tabular}

\end{table}

\begin{table}\footnotesize
\renewcommand{\arraystretch}{1.3}
	\caption{\small\textbf{Ablation studies} of \texttt{zero-shot} setting with additional text data on VOC 2007.}
	\label{tabzero}
        \setlength{\tabcolsep}{18pt}
	\fontsize{9}{9}\selectfont
	\centering
	\begin{tabular}{l  c c c }
\toprule
\textbf{Dataset}
&\multicolumn{1}{c}{T2I-PALw/$2$$\times$}
&\multicolumn{1}{c}{T2I-PAL}
\\
\cmidrule(lr){1-1} \cmidrule(l){2-3} 

{\cellcolor{myblue}VOC 2007}
&\multicolumn{1}{c}{{\cellcolor{myblue}\textbf{$91.8$}}}
&\multicolumn{1}{c}{{\cellcolor{myblue}\textbf{$91.5$}}}\\

\bottomrule
\end{tabular}

\end{table}

\begin{figure*}[t]
\centering  
\includegraphics[width=\linewidth]{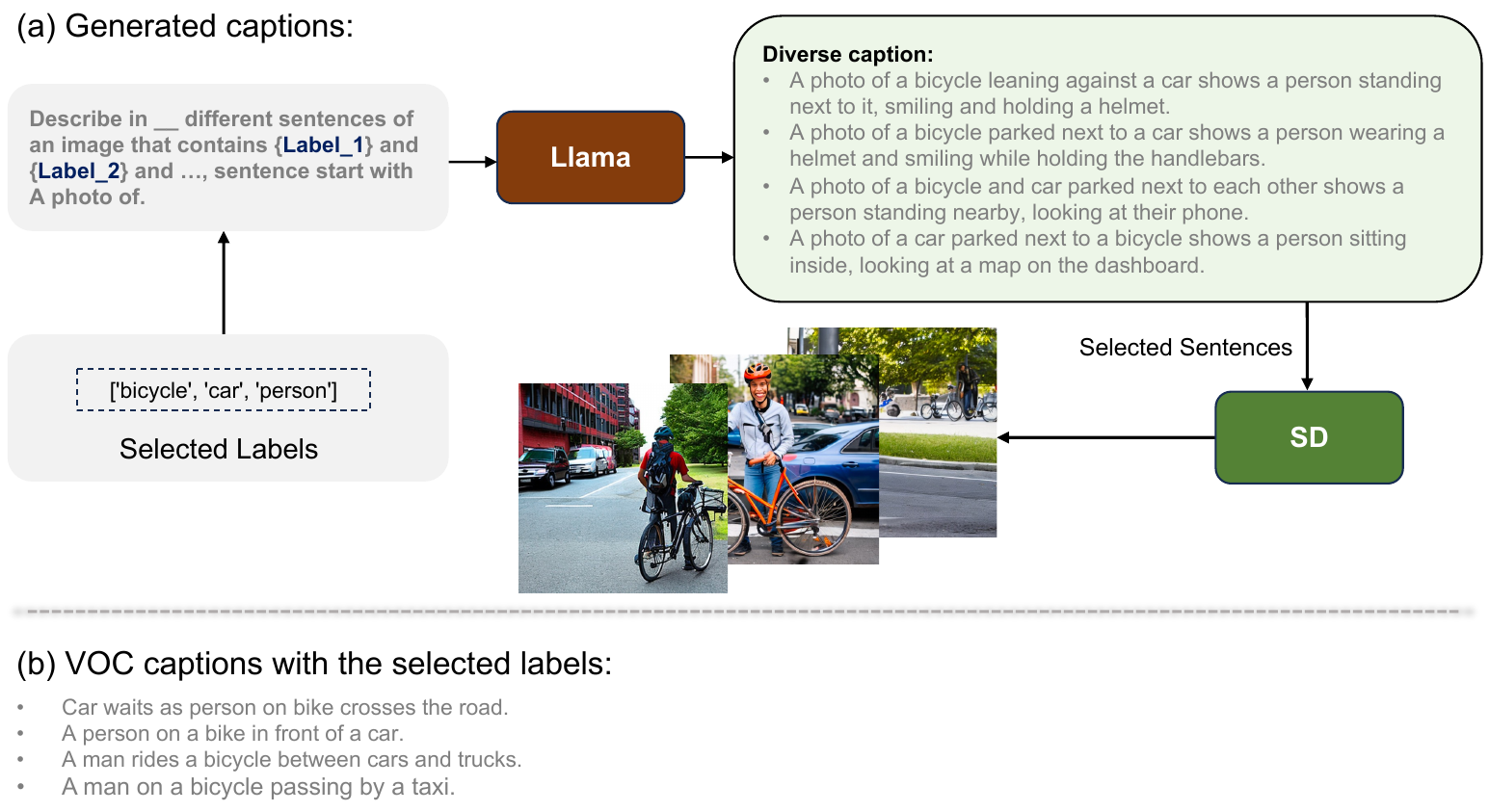}
\caption{(a) Illustration of leveraging Llama to generate text captions; (b) Examples of VOC captions with the selected labels. As a comparison, Llama can also provide quite complex and diverse text captions.}
\label{Fig-loss}
\end{figure*}

\begin{table*}[h]
\renewcommand{\arraystretch}{1.3}
	\caption{\small\textbf{Ablation studies} of \texttt{partial label} setting with additional text data on VOC 2007.}
	\label{tabpartail}
        \setlength{\tabcolsep}{11pt}
	\fontsize{9}{9}\selectfont
	\centering
	\begin{tabular}{l l  cc cc cc cc cc cc cc cc cc cc}
\toprule
&Method
&\multicolumn{2}{c}{\textbf{$10$\%}}
&\multicolumn{2}{c}{\textbf{$20$\%}}
&\multicolumn{2}{c}{\textbf{$30$\%}}
&\multicolumn{2}{c}{\textbf{$40$\%}}
&\multicolumn{2}{c}{\textbf{$50$\%}}
&\multicolumn{2}{c}{\textbf{$60$\%}}
&\multicolumn{2}{c}{\textbf{$70$\%}}
&\multicolumn{2}{c}{\textbf{$80$\%}}
&\multicolumn{2}{c}{\textbf{$90$\%}}
&\multicolumn{2}{c}{\textbf{Avg.}}
\\ 

\cmidrule(r){2-2}  
\cmidrule(r){3-20} 
\cmidrule(r){21-22} 

\multirow{3}{*}{\rotatebox{90}{\textbf{VOC}}}

&\multicolumn{1}{|l}{{+{T2I-PAL}} w/$2$$\times$}
&\multicolumn{2}{|c}{$93.9$} 
&\multicolumn{2}{c}{$94.8$}
&\multicolumn{2}{c}{$94.9$}  
&\multicolumn{2}{c}{$95.0$}
&\multicolumn{2}{c}{$94.9$}  
&\multicolumn{2}{c}{$95.5$}
&\multicolumn{2}{c}{$95.7$}  
&\multicolumn{2}{c}{$95.7$}
&\multicolumn{2}{c}{$95.7$}
&\multicolumn{2}{|c}{$95.1$}  \\ 

&\multicolumn{1}{|l}{\cellcolor{myblue}{{+{T2I-PAL(Ours)}}}} 
&\multicolumn{2}{|c}{{\cellcolor{myblue}\textbf{93.7}}} 
&\multicolumn{2}{c}{{\cellcolor{myblue}\textbf{94.8}}}
&\multicolumn{2}{c}{{\cellcolor{myblue}\textbf{94.8}}}
&\multicolumn{2}{c}{{\cellcolor{myblue}\textbf{94.9}}}
&\multicolumn{2}{c}{{\cellcolor{myblue}\textbf{94.9}}}
&\multicolumn{2}{c}{{\cellcolor{myblue}\textbf{95.2}}}
&\multicolumn{2}{c}{{\cellcolor{myblue}\textbf{95.5}}}
&\multicolumn{2}{c}{{\cellcolor{myblue}\textbf{95.5}}}
&\multicolumn{2}{c}{{\cellcolor{myblue}\textbf{95.5}}}
&\multicolumn{2}{|c}{{\cellcolor{myblue}\textbf{95.0}}}\\

\bottomrule
\end{tabular}
\end{table*}

\vspace{4pt}
\noindent\textbf{Modality Gap between Original and Synthetic Images.}
Albeit our T2I-PAL is effective, it is still difficult to fully eliminate the modality gap.
We provide the results of {\texttt{Ours}}({\texttt{O.I}}), {\texttt{Ours}} $1$$\times$ {\texttt{S.I}}, and T2I-PAL (\ie, {\texttt{Ours}} ($6$$\times${\texttt{S.I}})) under zero-shot setting, across three datasets, in Table~\ref{Amodality}.
As can be seen, {\texttt{Ours}} $1$$\times$ ({\texttt{S.I}}) performs on par with {\texttt{Ours}}({\texttt{O.I}}), indicating that SD is promising in generating photo-realistic images, and greatly minimize the modality gap. 
Furthermore, by using $6$$\times$ synthetic images, T2I-PAL outperforms {\texttt{Ours}}({\texttt{O.I}}) on all datasets, indicating that the diversity provided by more synthetic images compensates for the modality gap.
We also present results for cases where original images constitute $40$\% and $80$\% of the total dataset, with the remaining images being synthetic.
As shown in the table, the performance differences across different mixtures of original and synthetic images are minimal when compared to using only original images (labeled as {\texttt{Ours}}({\texttt{O.I}})) or an equal number of fully synthetic images (labeled as {\texttt{Ours}} $1$$\times$ ({\texttt{S.I}})).

\subsection{Ablation Study on the Quality of Different Modality Data}\label{ab2}
\vspace{2pt}
\noindent\textbf{Influence of the Quality of Text Captions.}
For {MS-COCO} and VOC 2007, we use the text captions provided in the datasets, while for {NUS-WIDE} we acquire the necessary captions from the OpenImages. Thus, the quality of the constructed text descriptions can be ensured in our experiments. Besides, one can generate text captions using existing language models, \eg, LLaMA~\cite{touvron2023llama}. In Fig.~\ref{Fig-loss}, we show the diversity of the generated text captions. In the Fig.~\ref{cvprtext}, we also show the results using the text captions generated by LLaMA~\cite{touvron2023llama}. One can see that \textbf{\texttt{Ours}}(\textbf{\texttt{Text.LLaMA}}) achieves similar performance with T2I-PAL, which shows its robustness \textit{w.r.t} the quality of the text description. Further, existing language models (\eg, LLaMA) have been effective in generating text descriptions with satisfying quality. 
Additionally, we also report the detailed results regarding the partial-label and few-shot settings.
Thus, in practice, the quality of the caption is not a major concern for our method.

\begin{table*}[b]
\renewcommand{\arraystretch}{1.7}
	\caption{\small \textbf{Comparison} with different visual encoders, \ie, ResNet$50$ and ResNet$101$, where ${\color{ForestGreen}\uparrow}$ and ${\color{red}\downarrow}$ indicates \textbf{improvements} and \textbf{decrements} compared with TaI-DPT~\cite{guo2022texts}.}
	\label{aptab1}
        \setlength{\tabcolsep}{20.9pt}
	\fontsize{9}{9}\selectfont
	\centering
	\begin{tabular}{c l  c c c}
\toprule
&Method
&\multicolumn{1}{c}{MS-COCO}
&\multicolumn{1}{c}{VOC 2007} 
&\multicolumn{1}{c}{NUS-WIDE}

\\ 

\cmidrule(r){2-2} \cmidrule(lr){3-3} \cmidrule(lr){4-4}\cmidrule(lr){5-5}
\multirow{2}{*}{\rotatebox{90}{\texttt{RN50}} }
&\multicolumn{1}{|l}{TaI-DPT${\color{gray}_{\text{[CVPR}23]}}$~\cite{guo2022texts}}  
&\multicolumn{1}{|c}{$88.30$} 
&\multicolumn{1}{c}{$65.10$}
&\multicolumn{1}{c}{$46.50$} 

\\

&\multicolumn{1}{|l}{\cellcolor{myblue}{{{T2I-PAL(Ours)}}}} 
&\multicolumn{1}{|c}{{\cellcolor{myblue}\textbf{88.80}\stdvu{$\underline{{0.5}}$}}}
&\multicolumn{1}{c}{{\cellcolor{myblue}\textbf{66.10}\stdvu{$\underline{{1.0}}$}}} 
&\multicolumn{1}{c}{{\cellcolor{myblue}\textbf{45.50}\stdvd{$\underline{{1.0}}$}}}\\ \hline\hline

\multirow{2}{*}{\rotatebox{90}{\texttt{RN101}}} 
&\multicolumn{1}{|l}{TaI-DPT${\color{gray}_{\text{[CVPR}23]}}$~\cite{guo2022texts}}  
&\multicolumn{1}{|c}{$88.30$} 
&\multicolumn{1}{c}{$65.40$}
&\multicolumn{1}{c}{$45.30$} 
\\

&\multicolumn{1}{|l}{\cellcolor{myblue}{{{T2I-PAL(Ours)}}}} 
&\multicolumn{1}{|c}{{\cellcolor{myblue}\textbf{91.50}}\stdvu{$\underline{{3.2}}$}}
&\multicolumn{1}{c}{{\cellcolor{myblue}\textbf{71.40}\stdvu{$\underline{{6.0}}$}}}
&\multicolumn{1}{c}{{\cellcolor{myblue}\textbf{47.40}\stdvu{$\underline{{2.1}}$}}}\\ 

\bottomrule
\end{tabular}
\end{table*}

\vspace{4pt}
\noindent\textbf{Influence of the Quality of Synthetic Images.}\label{sec:A2}
Since our method builds on the SD, the quality of synthetic images also affects the model performance. In general, better classification performance may be attained when synthetic images are of higher quality.
To illustrate this, we compare the performance using the synthetic images generated by SD $2.0$ (\ie, T2I-PAL) and SD $1.4$ (\ie, \textbf{\texttt{Ours}}(SD $1.4$)) in Fig.~\ref{cvprtext}. The results across the three datasets show that better image quality gives rise to slightly better MLR performance.
We try to use the compositional generation methods~\cite{liu2022compositional} to generate synthetic image. Under the zero-shot setting, our method improves the performance from $71.4$ to $71.8$ on MS-COCO dataset, while only improving $0.1$ on the VOC 2007.

\begin{figure}[t]
	\begin{center}
		\includegraphics[width=0.8\linewidth]{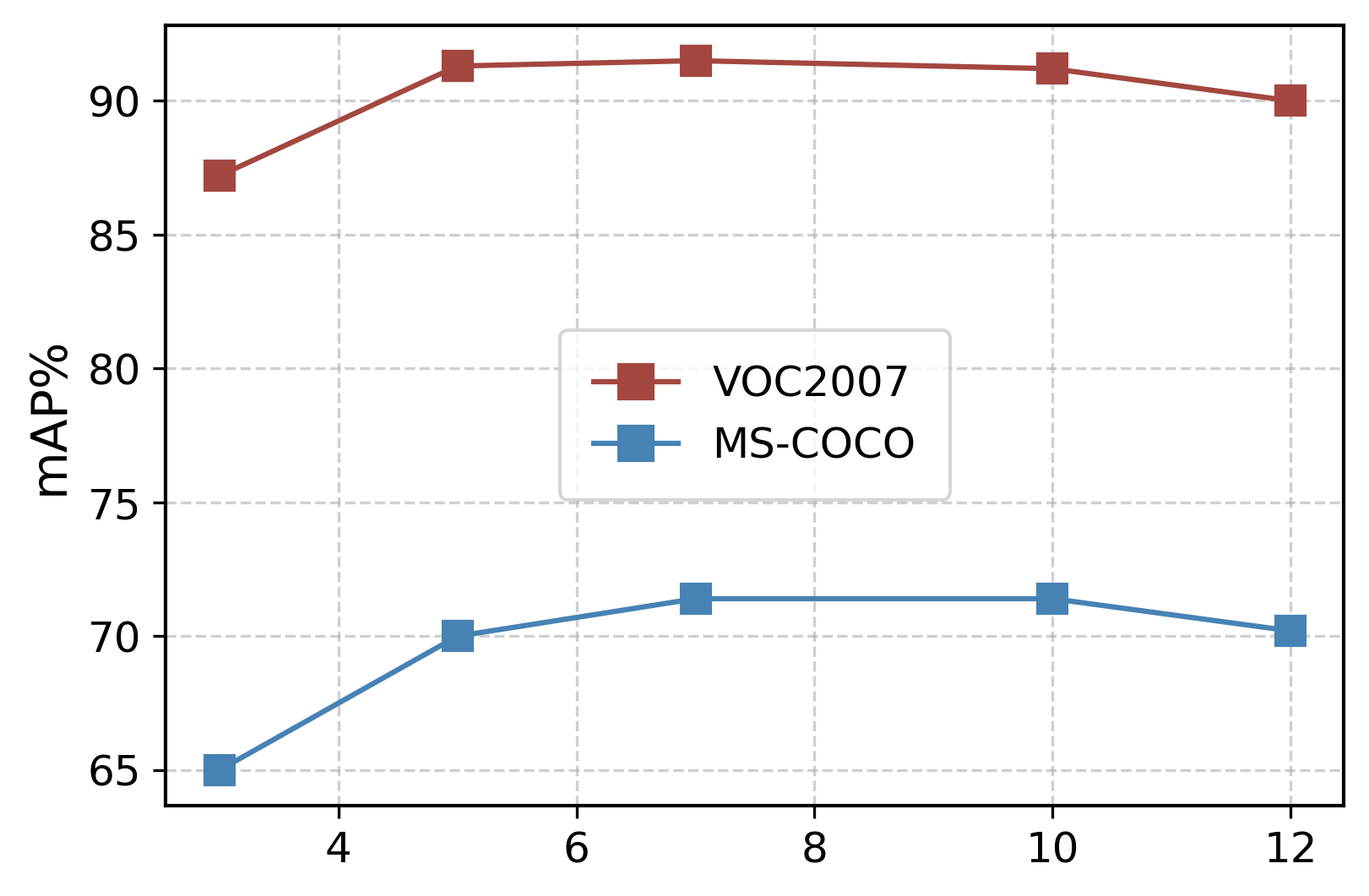}
	\end{center}
	\vspace{-8pt}
	\captionsetup{font=small}
    	\caption{Zero-shot performance of our method at different CFG scales.}
	\vspace{-13pt}
	\label{CFG}
\end{figure}

Besides, we also provide the zero-shot results using the SDXL~\cite{podell2023sdxlimprovinglatentdiffusion} as follows: $71.5$ on COCO, $91.7$ on VOC2007, and $47.5$ on NUS. We found that stronger T2I models lead to performance gains, as they often provide higher resolution outputs and more precise image details.
Additionally, we report the performance of our method under the zero-shot settings with different CFG scales in Fig.~\ref{CFG}. As shown in the figure, high CFG values can reduce the diversity of synthetic images, potentially leading to overfitting and a decrease in generalization ability. Conversely, lower CFG values may result in synthetic images that do not align well with the text prompt, leading to semantic inconsistencies and negatively affecting training, especially when the images do not match the target category. If synthetic images poorly match the text prompt, increasing the CFG value can improve classification performance. On the other hand, if the images become too similar due to overfitting, reducing the CFG value may enhance the diversity of the generated images and improve the model's generalization ability.

\begin{table*}[t]
\renewcommand{\arraystretch}{1.3}
	\caption{\small\textbf{Ablation studies} of \texttt{few-shot setting} with additional text data on VOC 2007.}
	\label{tabfew}
        \setlength{\tabcolsep}{14pt}
	\fontsize{9}{9}\selectfont
	\centering
	\begin{tabular}{c l  c c c c c c}
\toprule
&Method
&\multicolumn{1}{c}{\textbf{0-Shot}}
&\multicolumn{1}{c}{\textbf{1-Shot}} 
&\multicolumn{1}{c}{\textbf{2-Shot}}
&\multicolumn{1}{c}{\textbf{4-Shot}}
&\multicolumn{1}{c}{\textbf{8-Shot}}
&\multicolumn{1}{c}{\textbf{16-Shot}}
\\ 

\cmidrule(r){2-2} \cmidrule(lr){3-3} \cmidrule(lr){4-4}\cmidrule(lr){5-5}\cmidrule(lr){6-6}\cmidrule(lr){7-7}\cmidrule(lr){8-8}

&\multicolumn{1}{l}{{+{T2I-PAL}} w/$2$$\times$}
&\multicolumn{1}{|c}{$-$} 
&\multicolumn{1}{c}{$91.9$}
&\multicolumn{1}{c}{$92.2$} 
&\multicolumn{1}{c}{$92.3$} 
&\multicolumn{1}{c}{$92.5$}
&\multicolumn{1}{c}{$93.0$}
\\

&\multicolumn{1}{l}{\cellcolor{myblue}+{{{T2I-PAL(Ours)}}}} 
&\multicolumn{1}{|c}{{\cellcolor{myblue}$-$}}
&\multicolumn{1}{c}{{\cellcolor{myblue}\textbf{91.7}}}
&\multicolumn{1}{c}{{\cellcolor{myblue}\textbf{92.1}}}
&\multicolumn{1}{c}{{\cellcolor{myblue}\textbf{92.2}}} 
&\multicolumn{1}{c}{{\cellcolor{myblue}\textbf{92.3}}} 
&\multicolumn{1}{c}{{\cellcolor{myblue}\textbf{92.9}}}\\ 

\bottomrule
\end{tabular}
\end{table*}

\begin{figure*}[t]
	\begin{center}
		\includegraphics[width=\linewidth]{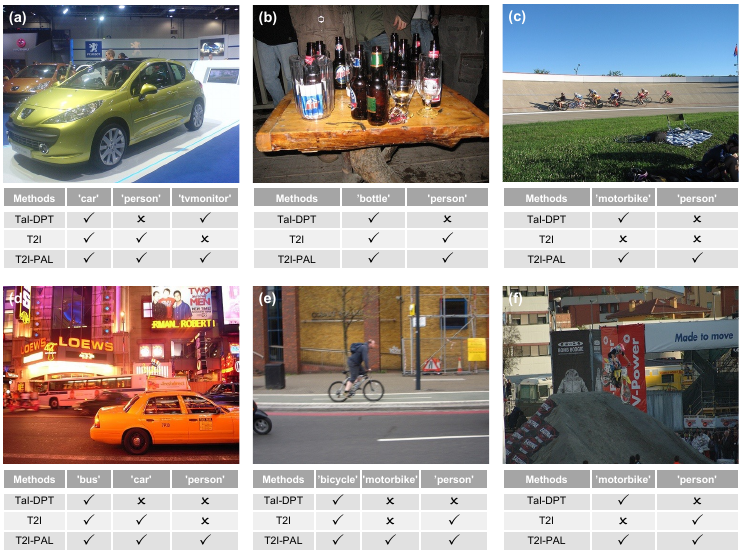}
	\end{center}
	\captionsetup{font=small}
	\caption{\textbf{Exemplars} of the complementarity of TaI-DPT~\cite{guo2022texts} and T2I, where \protect\includegraphics[scale=0.20,valign=c]{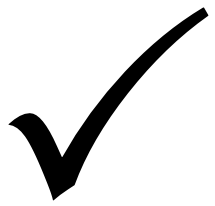} and \protect\includegraphics[scale=0.20,valign=c]{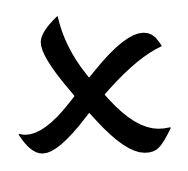} refer to the \textbf{correct} and \textbf{wrong} predictions of each class, respectively.}
	\label{figA1}
\end{figure*}

\vspace{4pt}
\noindent\textbf{Effect with Additional Text Data.}
We use LLaMA~\cite{touvron2023llama} to generate extra text captions containing one or several category names.
More details on leveraging LLaMA~\cite{touvron2023llama} to generate text captions are provided in Fig.~\ref{Fig-loss}. Then, we use Stable Diffusion to generate the corresponding synthetic images.
In this way, the size of text captions is doubled (\ie, T2I-PAL w/ $2$$\times$ text). Using VOC 2007, we provide the results of T2I-PAL and T2I-PAL w/ $2$$\times$ text in Table~\ref{tabzero}, ~\ref{tabpartail} and~\ref{tabfew}.
Benefiting from pre-trained language models, the generated text captions can be of both high quality and high diversity. 
One can see that, the introduction of extra text captions consistently improves classification performance under zero-shot, partial-label, and few-shot settings.

\subsection{Ablation Study on the Network Architecture}\label{ab3}
\vspace{4pt}
\noindent\textbf{Effect of PEFT.} %
As mentioned in the Section~\ref{adapter}, we share the adapter on the local features between the text caption and synthetic image branches to enhance the class representation of the model. Therefore, here we examine how the adapter affects the effectiveness of our method from three aspects, \ie, adapter on the global features, without adapter, and the hyperparameter analysis of the adapter. We first analyze whether using an adapter on global features can improve the performance of T2I-PAL. To this end, under the same hyperparameter setting with our T2I-PAL, we constructed two variants of T2I-PAL, \emph{w.} \texttt{Glo.Adp}, indicating that our method uses an adapter on global features, and \emph{w/o.} \texttt{Adp}, indicating that our method does not contain the adapter module. Table~\ref{tab6} records the performance of these variants on the three datasets. As can be seen from the table that \emph{w/o.} \texttt{Adp} has the worst performance on the three datasets.
%
%
This suggests that sharing an adapter between two modalities is helpful for enhancing MLR. Additionally, although the performance of \emph{w.} \texttt{Glo.Adp} has slightly improved, its performance is still far below that of our full model, T2I-PAL, \eg, \emph{w.} \texttt{Glo.Adp}: $70.7$ \textit{vs.} \texttt{Ours}: \textbf{71.4} on MS-COCO. This is attributed to the fact that more category knowledge can be captured on local features, thereby enhancing MLR.
Additionally, the number of adapter parameters is proportional to the number of categories: only $0.156$ MB for MSCOCO and $0.039$ MB for VOC2007. In contrast, the parameters for the foundation model, CLIP/RN50, amount to $85$ MB.
To this end, our full model, \emph{w.} \texttt{Loc.Adp} (\texttt{Ours}), shares an adapter module between local features of the two modalities, which can further tackle the modality gap issue when using only text captions for PEFT.
The hyperparameters closely related to the adapter in Eq.~(\ref{eq6}) and (\ref{eq7}), \ie, $\alpha$, the residual ratio of the features of the CLIP's text or visual encoder; and $\beta$, the modulating hyper-parameter that controls the sharpness of the affinities, are provided in Section~\ref{ab3}.

\vspace{4pt}
\noindent\textbf{Effect of Visual Encoders.}\label{sec:B}
Here, we investigate the impact of different visual encoders, \ie, ResNet$50$ and ResNet$101$, on model performance. As such, we use ResNet$50$ and ResNet$101$ as visual encoders for TaI-DPT and {T2I-PAL}, respectively. The zero-shot results on the three datasets are summarized in Table~\ref{aptab1}. As listed in this table, there is almost no difference in the performance of TaI-DPT on the different visual encoders, \eg, on the ResNet$50$, TaI-DPT~\cite{guo2022texts} obtain the performance of $88.3$ on VOC 2007, $65.1$ on MS-COCO, and $46.5$ on NUS-WIDE; on the ResNet$101$, TaI-DPT~\cite{guo2022texts} obtains the performance of $88.3$ on VOC 2007, $65.4$ on MS-COCO, and $45.3$ on NUS-WIDE. In particular, TaI-DPT~\cite{guo2022texts} decreases the performance on the NUS-WIDE from $46.5$ to $45.3$ when using ResNet$101$ as the visual encoder. On contrary, our method improves the performance on the three datasets when using ResNet$101$ as the visual encoder, \ie, on the ResNet$50$, {T2I-PAL} obtains the performance of $88.8$ on VOC 2007, $66.1$ on MS-COCO, and $45.5$ on NUS-WIDE; on the ResNet$101$, {T2I-PAL} obtains the performance of $91.5$ on VOC 2007, $71.4$ on MS-COCO, and $47.4$ on NUS-WIDE. More importantly, even based on ResNet$50$, our method still achieves improvements over TaI-DPT~\cite{guo2022texts}, \ie, $0.5$ gains on VOC 2007 and $1.0$ gains on MS-COCO. In summary, as TaI-DPT~\cite{guo2022texts} is based on text description, there is little difference in its performance for different visual encoders, while our method is based on synthetic images, making different visual encoders perform differently. Nevertheless, our method outperforms TaI-DPT~\cite{guo2022texts} on two datasets, \ie, MS-COCO and VOC 2007, no matter if it is based on ResNet$50$ or ResNet$101$.

\begin{figure*}
	\begin{center}
		\includegraphics[width=\linewidth]{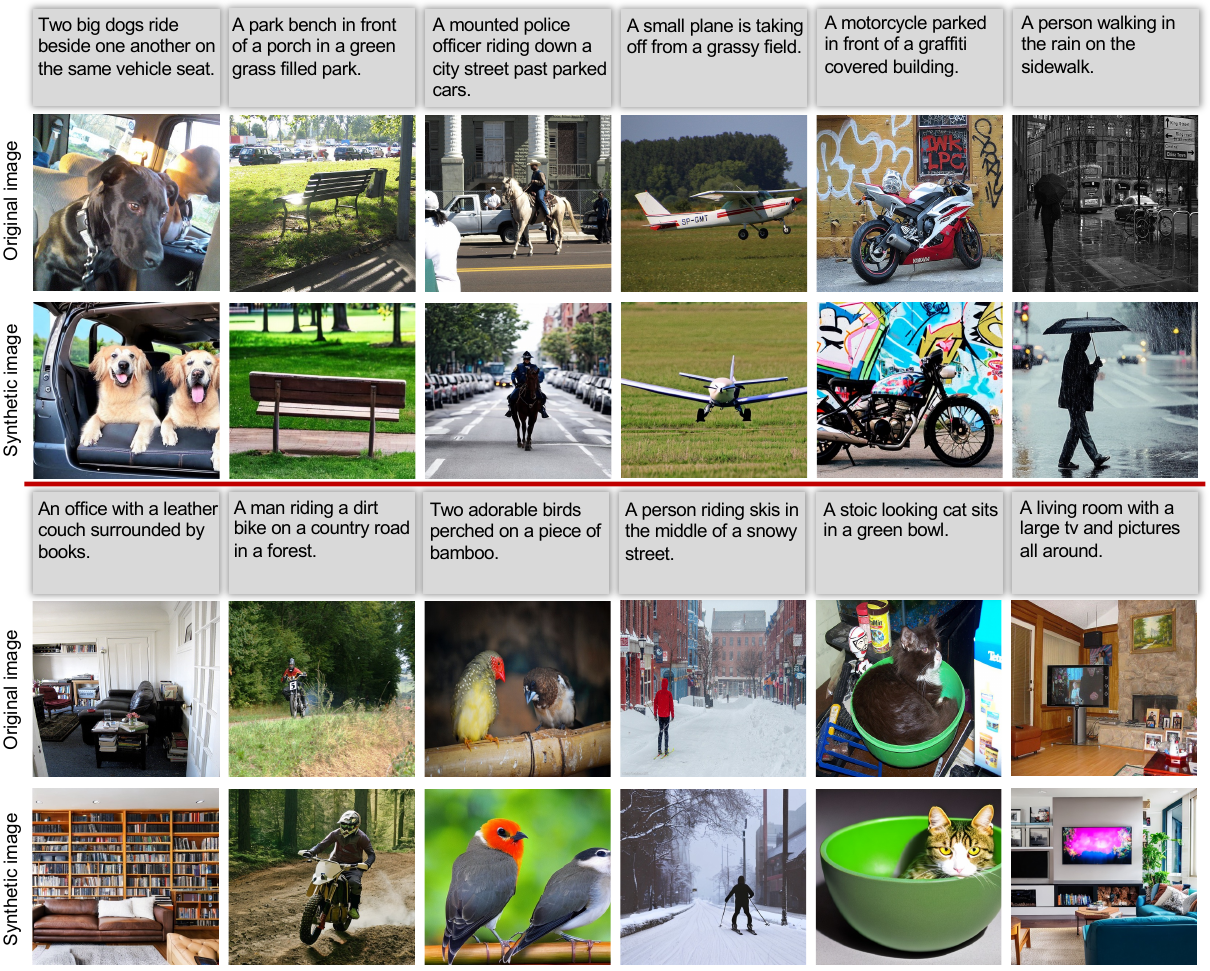}
	\end{center}
	\vspace{-8pt}
	\captionsetup{font=small}
    	\caption{\textbf{Visualization} of the \textbf{synthetic} image with their corresponding \textbf{text description} and \textbf{original} image.}
	\vspace{-13pt}
	\label{figa2}
\end{figure*}

\vspace{4pt}
\noindent\textbf{Complementary of TaI and T2I.}\label{sec:C}
As we have mentioned in Section~\ref{method} that our method absorbs the complementary merits of text captions and synthesized images from the pre-trained text-to-image generation model, we list some examples in Fig.~\ref{figA1} to illustrate it intuitively. This figure lists the predictions of each class in each corresponding image, where T2I refers to our method but without text description, \protect\includegraphics[scale=0.20,valign=c]{duihao.png} and \protect\includegraphics[scale=0.20,valign=c]{chahao.png} refer to the \textbf{correct} and \textbf{wrong} predictions of each class, respectively. As can be seen from this figure, TaI-DPT~\cite{guo2022texts} produces wrong predictions in some categories, while T2I produced wrong predictions in the remaining categories, \eg, in image \textbf{(a)}, TaI-DPT~\cite{guo2022texts} produces \textbf{wrong} predictions in the class of "\texttt{person}", while T2I produces \textbf{wrong} predictions in the class of "\texttt{tvmonitor}". Nonetheless, our proposed method, {T2I-PAL}, absorbs the complementary merits of text captions and synthesized images, thereby yielding correct predictions for these two categories, \ie, "\texttt{person}" and "\texttt{tvmonitor}". Interestingly, we also found that such complementary merits are effective in correcting predictions that are originally predicted incorrectly by both TaI-DPT~\cite{guo2022texts} and T2I, \eg, in image \textbf{(c)}, both TaI-DPT~\cite{guo2022texts} and T2I produce the \textbf{wrong} prediction in class "\texttt{person}", while {T2I-PAL} yields the correct prediction. These performances confirm that our method, which absorbs the complementary merits of both text captions and synthesized images, is an effective way for MLR.

\vspace{4pt}
\noindent\textbf{Visualization of the Synthetic Images.}\label{sec:D}
Here, we visualize some synthetic images, their corresponding original images, and text descriptions in Fig.~\ref{figa2}, where the synthetic images are generated by Stable Diffusion from the text descriptions. One can observe from the figure that the synthetic image generated from the text description preserves the key semantic information of the original image. In particular, the synthetic images are photo-realistic and diverse, providing an effective way for MLR while using only text captions. In contrast, TaI-DPT~\cite{guo2022texts} directly  for parameter-efficient fine-tuning, leading to a modality gap in different feature spaces.
Additionally, we present the failure cases of image generation in Fig.~\ref{failure}. Although SD can generate photo-realistic images, it may still lead to the omission of certain objects.

\begin{figure*}[t]
	\vspace{-9pt}
	\begin{center}
		\includegraphics[width=0.7\linewidth]{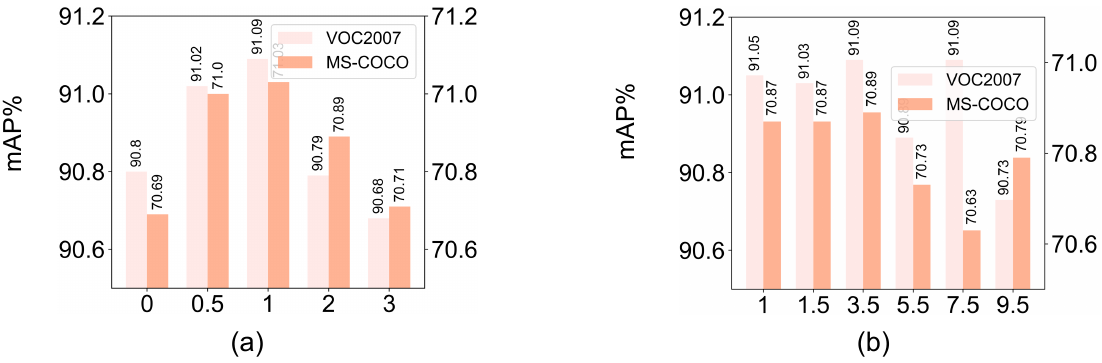}
        \put(-270,3){ \small{$\alpha$}}
        \put(-68,3){ \small{$\beta$}}
	\end{center}
	\vspace{-14pt}
	\captionsetup{font=small}
	\caption{\small\textbf{Analysis} with regard to the different values of $\alpha$, and $\beta$ on the two datasets, \ie, {VOC 2007} and MS-COCO, where \textbf{(a)} the \textit{smaller} the value of $\alpha$, the more \textit{prior knowledge} needs to be acquired from the pre-trained CLIP's visual encoder, vice versa; and \textbf{(b)} $\beta$ controls the \textit{sharpness} of the affinities.}
	\vspace{-10pt}
	\label{fig8new}
\end{figure*}

\begin{table*}
\renewcommand{\arraystretch}{1.3}
	\caption{\small\textbf{Ablation studies} of Eq.~(\ref{eq3}) and Eq.~(\ref{eq7}).}
\label{max}
\setlength{\tabcolsep}{11.5pt}
\fontsize{9}{9}\selectfont
\centering
\begin{tabular}{l  c c cc cc cc c}
\toprule

\textbf{{Variation}}
&\multicolumn{1}{c}{\textbf{\texttt{{Mean}}}}
&\multicolumn{1}{c}{\textbf{\texttt{{Max}}}}
&\multicolumn{2}{l}{\textbf{{~MS-COCO~}} }
&\multicolumn{2}{l}{\textbf{{~VOC-2007~}}}
&\multicolumn{2}{l}{\textbf{{~NUS-WIDE~}}}
&\multicolumn{1}{l}{\textbf{{~Average~}}}
\\
 \cmidrule(lr){1-1} \cmidrule(l){2-2} \cmidrule(l){3-3} \cmidrule(l){4-5} \cmidrule(l){6-7} \cmidrule(l){8-9} \cmidrule(l){10-10}
 
{T2I-PAL(Mean)}
&\multicolumn{1}{|c}{{\Checkmark}}
&$-$
&\multicolumn{2}{|l}{{$71.0$}\stdvd{{$\underline{0.4}$}}}
&\multicolumn{2}{l}{{$91.1$}\stdvd{{$\underline{0.4}$}}}
&\multicolumn{2}{l}{{$46.9$}\stdvd{{$\underline{0.4}$}}}
&\multicolumn{1}{|l}{{$69.66$}\stdvd{{$\underline{0.44}$}}}
\\

{T2I-PAL(Max)}
&\multicolumn{1}{|c}{$-$}
&\multicolumn{1}{c}{{\Checkmark}}
&\multicolumn{2}{|l}{{$70.8$}\stdvd{{$\underline{0.6}$}}}
&\multicolumn{2}{l}{{$91.0$}\stdvd{{$\underline{0.5}$}}}
&\multicolumn{2}{l}{{$47.0$}\stdvd{{$\underline{0.4}$}}}
&\multicolumn{1}{|l}{{$69.60$}\stdvd{{$\underline{0.50}$}}}\\

{TaI-DPT+Adapter}
&\multicolumn{1}{|c}{$-$}
&\multicolumn{1}{c}{$-$}
&\multicolumn{2}{|l}{{$65.6$}\stdvd{{$\underline{5.8}$}}}
&\multicolumn{2}{l}{{$88.7$}\stdvd{{$\underline{2.8}$}}}
&\multicolumn{2}{l}{{$46.9$}\stdvd{{$\underline{0.5}$}}}
&\multicolumn{1}{|l}{{$67.07$}\stdvd{{$\underline{3.03}$}}}\\

{T2I-PAL($\mathbf{W}$)}
&\multicolumn{1}{|c}{$-$}
&\multicolumn{1}{c}{$-$}
&\multicolumn{2}{|l}{{$71.9$}\stdvu{{$\underline{0.5}$}}}
&\multicolumn{2}{l}{{$91.7$}\stdvu{{$\underline{0.2}$}}}
&\multicolumn{2}{l}{{$47.8$}\stdvu{{$\underline{0.4}$}}}
&\multicolumn{1}{|l}{{$70.37$}\stdvu{{$\underline{0.37}$}}}\\

{T2I-PAL($\mathbf{M}$)}
&\multicolumn{1}{|c}{$-$}
&\multicolumn{1}{c}{$-$}
&\multicolumn{2}{|l}{{$71.5$}\stdvu{{$\underline{0.1}$}}}
&\multicolumn{2}{l}{{$91.8$}\stdvu{{$\underline{0.3}$}}}
&\multicolumn{2}{l}{{$47.7$}\stdvu{{$\underline{0.2}$}}}
&\multicolumn{1}{|l}{{$70.33$}\stdvu{{$\underline{0.20}$}}}\\

\bottomrule
\end{tabular}
\end{table*}


\begin{figure}
	\begin{center}
		\includegraphics[width=\linewidth]{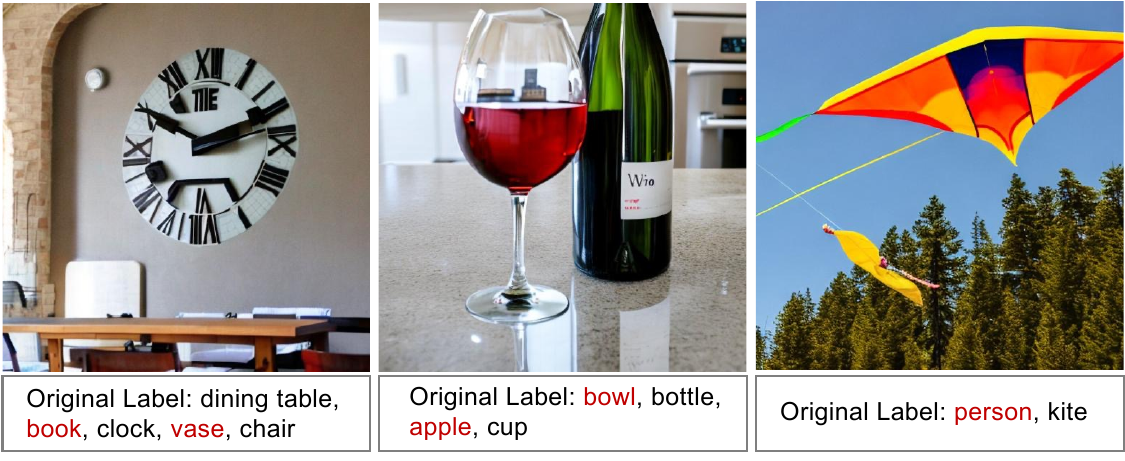}
	\end{center}
	\vspace{-8pt}
	\captionsetup{font=small}
    	\caption{{\textbf{Failure} cases of the synthetic image, where some objects are missing and marked in red.}}
	\vspace{-13pt}
	\label{failure}
\end{figure}

\vspace{4pt}
\noindent\textbf{Analysis of hyperparameters $\alpha$ and $\beta$.}\label{sec:E}
Here, we investigate two important hyperparameters closely related to the adapter in Eq.~(\ref{eq6}) and Eq.~(\ref{eq7}), \ie, $\alpha$, the residual ratio of the features of the CLIP's text or visual encoder; and $\beta$, the modulating hyper-parameter that controls the sharpness of the affinities. We plot the histogram of different values of $\alpha$ in Fig.~\ref{fig8new} (a), where the smaller the value of $\alpha$, the influence of prior knowledge from the pre-trained CLIP's visual encoder increases, while the larger the value of $\alpha$, the more knowledge needs to be learned from the adapter. As can be seen, when $\alpha$ = $0$, the model performs poorly because it degenerates into the zero-shot CLIP. As the value of $\alpha$ increases, the performance of the model starts to improve, with the highest results obtained at $\alpha$ = $1$, \ie, $91.09$ and $71.03$. However, when the value of $\alpha$ increases further, the performance of the model decreases, mainly because the prior knowledge from CLIP also plays an important role. That is, a good balance of knowledge learned from the adapter and prior knowledge that needs to be acquired from the pre-trained CLIP's visual encoder can enable the model to achieve the highest performance.
We then plot the histogram of different values of $\beta$ in Fig.~\ref{fig8new} (b). If the value of $\beta$ is large, then the classification prediction for a test image is primarily affected by the training samples in its vicinity, and vice versa. As can be seen from Fig.~\ref{fig8new} (b), when $\beta$ = $3.5$, our method achieves the highest performance. The results in Fig.~\ref{fig8new} (a) and (b) also show that the performance of different $\alpha$ and $\beta$ values on the different datasets is consistent. Given this, we set $\alpha$ = $1$, $\beta$ = $3.5$ in our experiments.

\noindent\textbf{Ablations of Eq.~(\ref{eq3}) and (\ref{eq7}).}
To further assess the effectiveness of Eq.~(\ref{eq3}), we simplify it by applying either the ‘mean’ or ‘max’ operation, with the results presented in Table~\ref{max}. As shown, the ‘mean’ operation simplifies the calculations but may lead to the loss of important local information, thereby reducing model performance. The ‘max’ operation highlights prominent areas but may neglect contributions from other regions, also resulting in decreased performance. In contrast, T2I-PAL aggregates local similarity with weighted contributions, accounting for both the significance of local areas and spatial distribution information, making it well-suited for multi-label image recognition tasks. We also include the results of TaI-DPT+Adapter in this table. The results show that with the adapter module, TaI-DPT+Adapter outperforms TaI-DPT, but still falls significantly short of our method, demonstrating the effectiveness of our core approach in MLR. Additionally, we modify Eq.~(\ref{eq3}) into a learnable attention block with a learnable parameter $\mathbf{W}$, as follows:
\begin{equation}
\mathbf{s}_i^{\prime}=\sum_{j=1}^{N_*} \frac{\exp \left(\mathbf{W} \cdot \mathbf{S}_{i j} / \tau\right)}{\sum_{j=1}^{N_*} \exp \left(\mathbf{W} \cdot \mathbf{S}_{i j} / \tau\right)} \cdot \mathbf{S}_{i j}.
\end{equation}
The results presented in T2I-PAL($\mathbf{W}$) in Table~\ref{max} demonstrate further improvement with the inclusion of a learnable parameter. A similar enhancement is observed in T2I-PAL($\mathbf{M}$), a variant of Eq.~(\ref{eq7}) that incorporates a learnable parameter $\mathbf{M}$:
\begin{equation}
\tilde{\mathbf{s}}_i^{\prime}=\mathbf{M}\left(\boldsymbol{q}_i, \mathbf{s}_i^{\prime}\right) \cdot \boldsymbol{q}_i+\left(1-\mathbf{M}\left(\boldsymbol{q}_i, \mathbf{s}_i^{\prime}\right)\right) \cdot \mathbf{s}_i^{\prime}.
\end{equation}
A learnable attention mechanism enables the model to automatically determine the optimal weight distribution, thereby reducing reliance on hyperparameters such as $\alpha$. While this approach can further enhance the model's classification performance and generalization ability, it also increases computational complexity. The increased computational complexity can lead to longer training times.

\section{Conclusion}
This paper presented a new PEFT method, T2I-PAL, based on a large-scale pre-trained vision-language model to address the modality gap issue when performing PEFT only with text captions. The core design of T2I-PAL is to utilize a pre-trained text-to-image generation model to synthesize photo-realistic and diverse images from text captions. T2I-PAL provides two appealing benefits: 1) it does not require any full semantically annotated training image, thereby lowering the burden of manual annotation; 2) it does not destroy the inherent mode of the CLIP model and can be implanted into any CLIP model. Additionally, T2I-PAL combines both prompt tuning and adapter learning with the two modalities, thereby enhancing classification performance.
Extensive experiments on multiple benchmarks such as MS-COCO, VOC 2007 and NUS-WIDE show that our T2I-PAL recognition performance significantly outperforms the top-ranked state-of-the-art methods. 
Potential limitation of this work is that our paper relies on pre-trained text-to-image generation models to generate photo-realistic and diverse images from text captions. However, it does not address potential limitations or challenges associated with these models, such as biases in the generated images or their capability to faithfully represent the intended visual content.


\section*{Acknowledgments}
This work was supported by the Agency for Science, Technology, and Research (A*STAR) through its AME Programmatic Funding Scheme Under Project A20H4b0141, the National Research Foundation (NRF) Singapore under its AI Singapore Programme (AISG Award No: AISG2-TC-2021-003), the Agency for Science, Technology, and Research (A*STAR) through its RIE2020 Health and Biomedical Sciences (HBMS) Industry Alignment Fund Pre-Positioning (IAF-PP) (grant no. H20C6a0032), and partially supported by A*STAR Central Research Fund "A Secure and Privacy-Preserving AI Platform for Digital Health”.

\bibliographystyle{ieee_fullname}
\bibliography{reference}

\end{document}